\documentclass[journal]{IEEEtran}

\usepackage{amssymb}
\usepackage{cite}
\usepackage{amsmath,amssymb}
\usepackage{algorithmic}
\usepackage{algorithm}
\usepackage{flushend}
\usepackage{array}
\usepackage{cuted}
\usepackage[utf8]{inputenc} 
\usepackage[T1]{fontenc}    
\usepackage{hyperref}       
\usepackage{url}            
\usepackage{booktabs}       
\usepackage{amsfonts}       
\usepackage{nicefrac}       
\usepackage{microtype}      
\usepackage{graphicx}
\usepackage{multirow}
\usepackage{colortbl}
\usepackage[normalem]{ulem}
\usepackage[justification=centering]{caption}
\usepackage{textcomp}
\usepackage{stfloats}
\usepackage{verbatim}
\usepackage{color}
\usepackage{enumerate}
\usepackage{arydshln}
\usepackage[table,xcdraw]{xcolor}
\newcommand{\orcid}[1]{\href{https://orcid.org/#1}{\includegraphics[width=10pt]{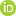}}}

\begin{document}
\title{SFNet: Fusion of Spatial and Frequency-Domain Features for Remote Sensing Image Forgery Detection}

\author{Ji Qi\orcid{0000-0001-7948-579X},
 Xinchang Zhang\orcid{0000-0002-3773-4898},
 Dingqi Ye\orcid{0000-0003-2260-7551},
 Yongjian Ruan,
 Xin Guo,
 Shaowen Wang\orcid{0000-0001-5848-590X},
 Haifeng Li\orcid{0000-0003-1173-6593}


\thanks{

This work was supported by the National Natural Science Foundation of China under Grants 42371406 and 42071441. \textit{(Corresponding author: D.Q. Ye)}

J. Qi, X.C. Zhang and Y.J. Ruan are with the School of Geography and Remote Sensing, Guangzhou University, Guangzhou 510006, China (e-mail: jameschi95@foxmail.com; zhangxc@gzhu.edu.cn; ruanyj@gzhu.edu.cn). X.C. Zhang is also with the Xiamen University Of Technology, Xiamen, 361024, Fujian Province, China and the College of Geography and Remote Sensing Sciences, Xinjiang University, Urumqi 830017, China;

D.Q. Ye and S.W. Wang are with Department of Geography and Geographic Information Science, University of Illinois at Urbana–Champaign, Urbana, IL 61820 USA, and also with the CyberGIS Center for Advanced Digital and Spatial Studies, University of Illinois at Urbana–Champaign, Urbana, IL 61801 USA (e-mail: dingqi2@illinois.edu; shaowen@illinois.edu).

X. Guo, H.F. Li are with the School of Geosciences and Info-Physics, Central South University, Changsha 410083, China. (e-mail: 225011031@csu.edu.cn, lihaifeng@csu.edu.cn).

}
}

\markboth{Journal of \LaTeX\ Class Files,~Vol.~14, No.~8, Apr~2025}%
{Shell \MakeLowercase{\textit{et al.}}: Bare Demo of IEEEtran.cls for IEEE Journals}

\maketitle

\begin{abstract}
  The rapid advancement of generative artificial intelligence is producing fake remote sensing imagery (RSI) that is increasingly difficult to detect, potentially leading to erroneous intelligence, fake news, and even conspiracy theories. Existing forgery detection methods typically rely on single visual features to capture predefined artifacts, such as spatial-domain cues to detect forged objects like roads or buildings in RSI, or frequency-domain features to identify artifacts from up-sampling operations in adversarial generative networks (GANs). However, the nature of artifacts can significantly differ depending on geographic terrain, land cover types, or specific features within the RSI. Moreover, these complex artifacts evolve as generative models become more sophisticated. In short, over-reliance on a single visual cue makes existing forgery detectors struggle to generalize across diverse remote sensing data. This paper proposed a novel forgery detection framework called SFNet, designed to identify fake images in diverse remote sensing data by leveraging spatial and frequency domain features. Specifically, to obtain rich and comprehensive visual information, SFNet employs two independent feature extractors to capture spatial and frequency domain features from input RSIs. To fully utilize the complementary domain features, the domain feature mapping module and the hybrid domain feature refinement module(CBAM attention) of SFNet are designed to successively align and fuse the multi-domain features while suppressing redundant information. Experiments on three datasets show that SFNet achieves an accuracy improvement of 4\%-15.18\% over the state-of-the-art RS forgery detection methods and exhibits robust generalization capabilities. The code is available at https://github.com/GeoX-Lab/RSTI/tree/main/SFNet.
\end{abstract}


\section{Introduction}
\IEEEPARstart{F}ake remote sensing imagery (RSI) is becoming increasingly difficult to detect as generative artificial intelligence (AI) models make significant progress \cite{Epstein_Hertzmann_2023, Zhang_Zhang_Zhang_Kweon_2023, Cao_Tan_Gao_Xu_Chen_Heng_Li_2024, OpenAI_2024}.
Forged RSI may be used to conceal military operations or constructions, spread misleading information, or create rumors\footnote{\href{https://www.forbes.com/sites/mattnovak/2023/03/27/ai-creates-photo-evidence-of-2001-earthquake-that-never-happened}{AI Creates Photo Evidence Of 2001 Earthquake That Never Happened}} \cite{temirDeepfakeNewEra2020, Zhao_Zhang_Xu_Sun_Deng_2021}.
The increasing prevalence of forged RSIs may not only lead to various social problems \cite{ciftciDeepfakeSatelliteImagery2023}, but also fundamentally undermine the credibility of authentic remote sensing data \cite{bennettBringingSatellitesEarth2024,albanwan2024image}.
Furthermore, the potential risks associated with forged RSIs will continue to escalate as generative AI technology advances.
However, detecting forged RSIs remains highly challenging, and progress has been slow \cite{Yarlagadda_Guera_Bestagini_Zhu_Tubaro_Delp_2018, Kuznetsov_2020, Zhao_Zhang_Xu_Sun_Deng_2021,lang2024spatial}.
In particular, the characteristics of RSIs, such as wide coverage and long imaging distances \cite{Cui_Zhang_Wang_Li_Qi_2021}, present the following main challenges:

Firstly, under the combined influence of atmospheric interference, radiation offset, and complex pre-processing, real remote sensing images (RSI) are also prone to typical "artifacts". For example, ortho-rectified RSI are prone to local distortions and shadows, the former being visually similar to common spatial-domain artifacts, while the latter are identical to typical frequency-domain artifacts. In addition, some image-based methods regard image content blurring or local noise as typical artifact representation, which is quite common in real RSI at long distances. Therefore, for the forgery detection of RSI, it is necessary to discard some of the inherent prejudices about artifact representation and re-examine the differences between real and forged RSI.

Secondly, remote sensing images are characterized by richness, variability and diversification of falsification techniques, which make it difficult to presuppose a generalized artifacts representation for detection. RSI can cover homogeneous scenes with uniform texture and color features (e.g., deserts and oceans) or heterogeneous scenes with multiple ground objects (e.g., residential areas and parking lots). Different scene types can be forged in various ways, which poses a greater challenge for designing robust forgery detection algorithms. In addition, due to RSI's spatial, temporal, and spectral heterogeneity, even the same object or scene often has different visual representations in different images. Therefore, it is difficult to presuppose an artifact representation or pattern to support forgery detection for various RSI.

In brief, RSI forgery detection requires advanced capabilities to capture and distinguish forgery patterns effectively. Existing forgery detection methods often show limitations when applied to RSI due to their reliance on specific forgery methods and predefined feature patterns, which are not adaptable to the complex and varied nature of RSI. Specifically, early approaches relied on manual feature extraction to detect specific artifacts in the spatial domain \cite{Agarwal_Farid_Gu_He_Nagano_Li_2019, Yang_Li_Lyu_2019, passosReviewDeepLearningbased2024}. However, these methods were only effective for low-quality forgeries that exhibit noticeable distortions and noise.

Subsequently, many researchers leveraged convolutional neural networks (CNNs) with powerful spatial feature extraction capabilities to automatically capture artifacts in forged images \cite{Sengur_Akhtar_Akbulut_Ekici_Budak_2018, moFakeFacesIdentification2018, Li_Lyu_2019, S_Thillaiarasu_2022}.

However, research has demonstrated that these methods may overfit to specific color and texture patterns associated with particular forgery models, limiting their ability to generalize to forged images created by other models \cite{Luo_Zhang_Yan_Liu_2021}.

Furthermore, with advancements in adversarial generative networks (GANs) \cite{Karras_Aila_Laine_Lehtinen_2018,Karras_Laine_Aila_2019,Thies_Zollhofer_Stamminger_Theobalt_Niessner_2016,Perov_Gao_Chervoniy_Liu_Marangonda_Ume_Dpfks_Facenheim_RP_Jiang_et}, artifacts in generated images have become increasingly subtle, making detection in the spatial domain significantly more challenging.
To address this challenge, many studies have shifted to capturing artifacts in the frequency domain, achieving impressive results \cite{Qian_Yin_Sheng_Chen_Shao_2020,Qin_Zhang_Wu_Li_2021,Tan_Zhao_Wei_Gu_Liu_Wei_2024,Tian_Luo_Shi_Li_2023}.

Recent major breakthroughs in generative AI, especially in diffusion models \cite{Ho_Jain_Abbeel_2020,Ramesh_Pavlov_Goh_Gray_Voss_Radford_Chen_Sutskever_2021,Rombach_Blattmann_Lorenz_Esser_Ommer_2022}, have significantly challenged the reliability of advanced detection methods that are based on frequency domain priors.
In general, the existing methods often target specific types of forged data and face significant limitations when applied to the diverse and evolving nature of RSI forgeries.

\begin{figure*}[]
  \centering
  \includegraphics[width=0.81\linewidth]{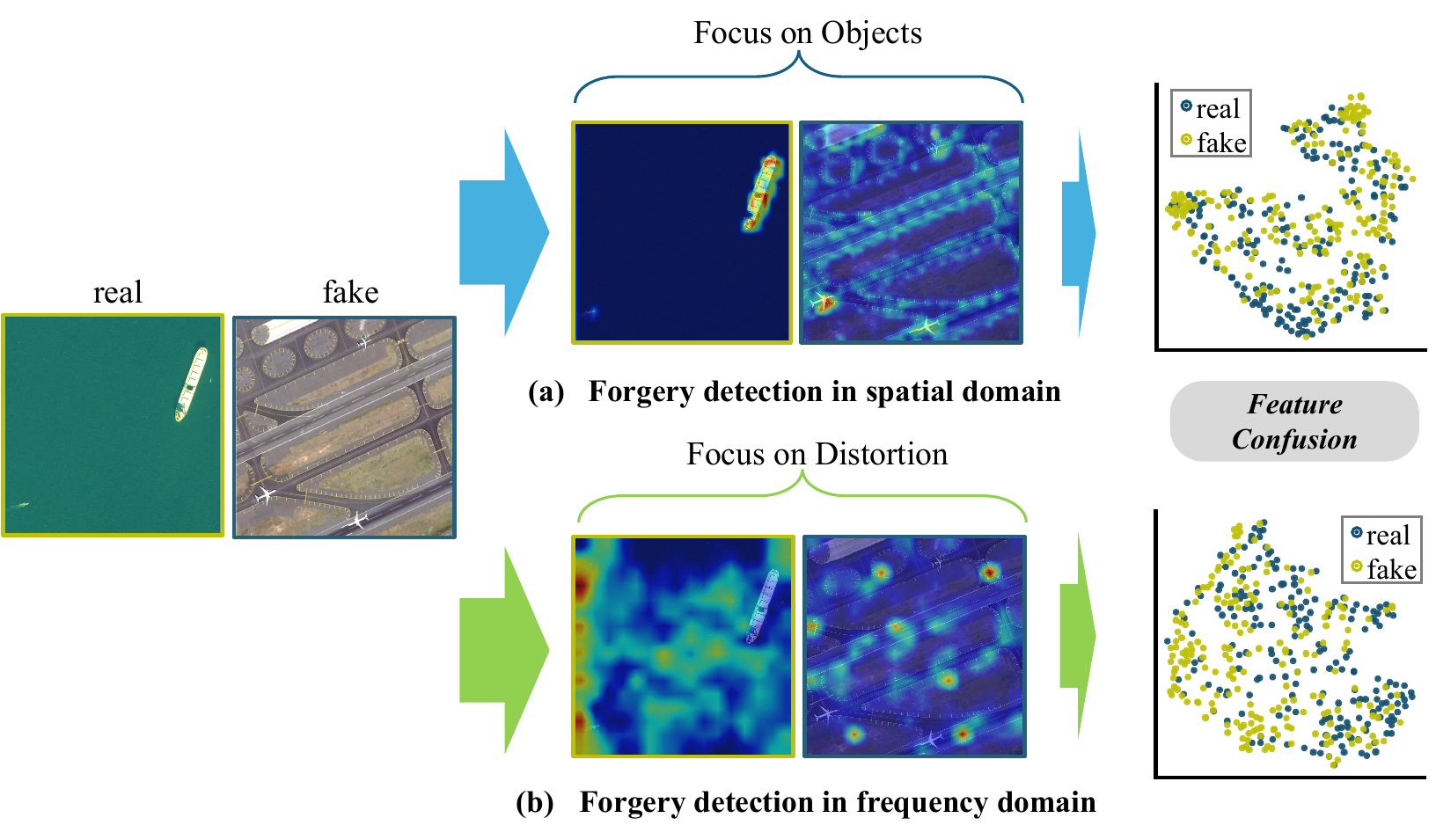}
  \caption{
    Problems with single-domain detection models in Remote sensing images (Frequency domain/Spatial domain).
  }
  \label{fig:intro}
\end{figure*}

To address the limitations of forgery detection methods that rely on single features and specific artifacts, this paper aims to improve forgery detection for complex and diverse RSIs by exploiting and fusing visual features from multiple domains. The problem with the current detection approach in remote sensing is shown in Fig.\ref{fig:intro}. We use GradCAM\cite{Selvaraju_Cogswell_Das_Vedantam_Parikh_Batra_2020} visualize which parts of an image contribute the most to a model's decision result, and t-SNE(t-Distributed Stochastic Neighbor Embedding)\cite{van2008visualizing} to visualize the model's ability to differentiate between true and fake image. (a) Due to subtle visual differences between real and fake images, the spatial feature-based model focuses on prominent objects (e.g., ships, roads, buildings) as shown in Grad-CAM visualization. But these objects appear in both real and fake images, making differentiation challenging, as seen in overlapping t-SNE results. (b) Using frequency feature, the model captures patterns not directly perceivable by the human eye. The real and fake images exhibit some separability in the t-SNE results, but significant overlap remains, indicating imperfect detection.

To achieve this goal, we propose the SFNet framework, consisting of four well-designed modules that jointly leverage spatial and frequency domain features for RSI forgery detection.
Specifically, spatial and frequency feature extractors are employed to capture rich and comprehensive visual information from the input RSI.
To fully utilize the complementarity of these domain features, the domain mapping module and the mix-domain feature refine module are designed to fuse the multi-domain features while suppressing redundant information.
With these modules, SFNet can more accurately detect artifacts in forged RSIs by dynamically focusing on distinguishable regions based on image content.
Experiments demonstrate that SFNet outperforms state-of-the-art methods in both accuracy and robustness on all three datasets.
Our main contributions are as follows:

\begin{itemize}
  \item We propose SFNet, a novel forgery detection model designed to overcome the limitations of existing methods that rely on single visual features to capture specific artifacts.
  With a series of specially designed modules, including two domain-specific feature extractors, a domain mapping module and a mix-domain feature refine module, SFNet effectively fuses spatial and frequency domain features to capture artifacts in diverse fake RSIs.

  \item To further verify the generalization ability of the model, we constructed an RSI forgery detection benchmark, SDGen-Detection, using an advanced stable diffusion model to support future research in remote sensing field.

  \item In experiments, SFNet outperforms the latest remote sensing forgery detection methods by 15.8\% in detection accuracy and demonstrates stable generalization ability. We conducted ablation studies to assess the contribution of each module. We also discussed dataset characteristics in spatial and frequency domains, the impact of data augmentation and pretrained models on performance, and suggested future research directions.
\end{itemize}

\section{Related Works}
\subsection{Spatial-based method}

Spatial-based forgery detection methods mainly exploit spectral, texture, geometry, and other spatial information to capture artifacts in AI-generated images and videos.

With powerful spatial feature extraction capability, CNNs such as AlexNet \cite{Krizhevsky_Sutskever_Hinton_2012} and VGG \cite{Simonyan_Zisserman_2014} are widely used in forgery detection tasks\cite{Sengur_Akhtar_Akbulut_Ekici_Budak_2018, moFakeFacesIdentification2018, Li_Lyu_2019}. To improve forgery detection accuracy, early studies introduced specific designs in detection models based on expert prior knowledge of common artifacts in the data. For example, Li and Lyu (2019) \cite{Li_Lyu_2019} found that affine transformations usually leave specific artifacts in forged face videos, and therefore designed specialized CNNs to distinguish real and fake data by targeting artifacts in affine face warping as distinctive features.

However, it is obvious that these methods based on expert priors are not applicable to RSIs. Moreover, objects and scenes in various RSIs with imaging conditions reveal infinite variety of spatial visual features, resulting in a lack of generalisation of a priori knowledge observed and summarised for specific targets and scenes. In this regard, \c Cift\c ci and Demir \cite{ciftciDeepfakeSatelliteImagery2023} observed and summarized a large number of RSIs generated by the GAN model, finding that texture and geometric distortions are common, and that target edges are often blurred. Based on these observations, they introduced a super-resolution enhancement technique and a multi-scale attention mechanism to capture specific artifacts in RSIs.

Nevertheles, there are numerous ways to manipulate RSIs, making it infeasible to cover all possible cases with limited and fixed expert prior knowledge \cite{Khodabakhsh_Ramachandra_Raja_Wasnik_Busch_2018}. To enhance the generalizability of forgery detection methods, some scholars try to improve the feature extraction capability of detection models for better adapting to complex and variable data and diverse forgery techniques \cite{chaiWhatMakesFake2020}. For example, Gowda and Thillaiarasu (2022) \cite{S_Thillaiarasu_2022} propose a model integration approach that provides complementary features for capturing artifacts by combining CNNs with different structures and characteristics, such as Xception \cite{Chollet_2017},ResNext \cite{Xie_Girshick_Dollar_Tu_He_2017} and Wodajo and Atnafu\cite{Wodajo_Atnafu_2021}.

In general, with the advancement of generative AI models, detecting forged images using spatial domain features has become increasingly challenging. The main reason is that generative models (e.g., Autoencoder, GAN, diffusion models) are typically trained by minimising the difference between the generated image and the real image in the spatial domain \cite{durallWatchYourConvolution2020, rautGenerativeAIVision2024}, making artifacts naturally difficult to capture using spatial domain features.

\subsection{Frequency-based method}

Pioneering studies have shown that frequency-domain features provide a unique and important perspective for representing and detecting artifacts in forged images \cite{huangRobustnessCopymoveForgery2017, zhangDetectingSimulatingArtifacts2019, durallWatchYourConvolution2020,Frank_Eisenhofer_Schonherr_Fischer_Kolossa_Holz_2020}. For example, Huang et al. \cite{huangRobustnessCopymoveForgery2017} use Fast Fourier Transform (FFT) and Singular Value Decomposition to extract global frequency-domain features from images that are insensitive to visual perturbations, achieving accurate recognition of copy-move manipulations under various noise and blurring attacks.

For AI-generated forged images, Frank et al. \cite{Frank_Eisenhofer_Schonherr_Fischer_Kolossa_Holz_2020} showed that GAN-generated images exhibit specific artifacts that are easily identified in the frequency domain. Durall et al. \cite{durallWatchYourConvolution2020} further investigated the cause of such frequency domain artifacts, suggesting that the typical up-sampling operations in GAN decoders changes the spectral properties of image \cite{jainFundamentalsDigitalImage1989}, distorting high-frequency information in the generated results.

However, different GAN models vary in network structure, data processing, training data, etc., resulting in fundamentally different frequency-domain artifacts in their generated results \cite{Frank_Eisenhofer_Schonherr_Fischer_Kolossa_Holz_2020}. Therefore, even well-trained detection models often perform well only on specific GAN-generated results, while the specific GAN model used by an attacker is usually unknown. To address the above problems, scholars have conducted research mainly from the perspectives of data and models to improve the generalization of forgery detection methods based on frequency-domain features.

First, from a data-based perspective, a representative approach is to provide the detector with diverse forgery images generated by a variety of models to help it learn robust features effectively. For instance, Zhang et al. \cite{zhangDetectingSimulatingArtifacts2019} proposed a GAN simulator, AutoGAN, to simulate several popular GAN models of forged images for training detection models. But this approach requires significant computational costs and longer training periods, and may not adapt to the rapid iteration of generative models.

Second, from a model-based perspective, researchers improve the robustness and adaptability of detection models by enhancing the mechanisms for extracting and utilizing frequency domain information. Qian et al. \cite{Qian_Yin_Sheng_Chen_Shao_2020} combined global and local frequency features to obtain richer artifact-aware clues for forgery detection. Jeong et al. \cite{Jeong_Kim_Min_Joe_Gwon_Choi_2022} proposed Bilateral High-Pass Filters (BiHPF) to amplify the frequency domain artifacts in GAN-generated images for generalized detection. In addition, some scholars introduced frequency-domain attention mechanisms to encourage the detector to prioritize critical regions of the image and suppress overfitting to low-level spatial features, thereby improving generalization \cite{jiaInconsistencyAwareWaveletDualBranch2021, singhalFrequencySpectrumMultihead2023, Binh_Woo_2022, Tan_Zhao_Wei_Gu_Liu_Wei_2024}.

In forgery detection tasks, overreliance on a single feature undoubtedly increases the risk of the model being spoofing. Research has shown that focusing too much on frequency domain artifacts may improve detection models' performance on training data, but can ultimately hinder its generalization to more diverse test datasets\cite{Jeong_Kim_Ro_Choi_2022}.

\subsection{Mixed-based method}

Recently, some studies have focused on overfitting forgery feature learning in a single domain. CNN detector was used to obtain specific color and texture information, and high-frequency feature detection module was used to extract high-frequency noise guidance features for feature complementation \cite{Luo_Zhang_Yan_Liu_2021}. Liu et al \cite{Liu_Li_Zhou_Chen_He_Xue_Zhang_Yu_2021} combines the features of  frequency spatial domains in the shallow layer, reducing the obvious changes in the cumulative up sampling frequency domain and improving the capability of the detection method. Chen et al \cite{Chen_Yao_Chen_Ding_Li_Ji_2021} proposed the RGB-Frequency Attention Module (RFAM), which fused the RGB and frequency domain information to achieve a more comprehensive local feature representation, and further improved the reliability of similar models. Jia et al 2023 \cite{Jia_Yao_2023} designed frequency domain and spatial domain learning to effectively capture intrinsic global and local information respectively. However, since the features of RGB and Frequency may have significant inconsistencies at the semantic level and at different scales, traditional additions or concatenations may adversely affect the fusion weights and limit the performance of forgery detection.

\subsection{RSI forgery detection}

Recently, more studies have begun to pay attention to satellite gaze and its policy implications. Zhao et al. \cite{Zhao_Zhang_Xu_Sun_Deng_2021} emphasized the importance of the problem and proposed the first remote sensing false image detection data set and monitoring method based on cycled GAN in the field of remote sensing. Yarlagadda et al. \cite{Yarlagadda_Guera_Bestagini_Zhu_Tubaro_Delp_2018} highlighted the difference between consumer cameras and overhead remote sensing images. Using generative adversarial networks (GAN), feature representations of raw satellite images can be learned. In addition, Patil et al. \cite{Patil_Chaudhari_Narawade_2022} proposed a variant of U-net and performed well on the RSICD dataset utilizing forgery techniques such as copy-move and splicing. Chen et al. \cite{Chen_Zhang_Hu_You_Kuo_2021} proposed a monitoring framework Geo-DefakeHop based on the parallel subspace learning (PSL) method. PSL uses multiple filter banks to map the input image space into multiple feature subspaces, learns the most discriminative channels by exploring the response differences of different channels between real and fake images, and uses their soft decision scores as features. Overall, the field of remote sensing lacks open source data sets and reproducible methods for AI-based forgery detection research.

\section{Methodology}
\label{methodology}

\subsection{Overview}

The overall framework of SFNet is shown in Figure \ref{fig:overview}. For the input image, we extract the spatial and frequency features in the shallow layer, then we fuse the aligned features through the projection layer for further deep feature extraction, and finally output the detection results through the linear layer. There are two key designs:

\begin{figure*}[]
    \centering
  \includegraphics[width=0.9\linewidth]{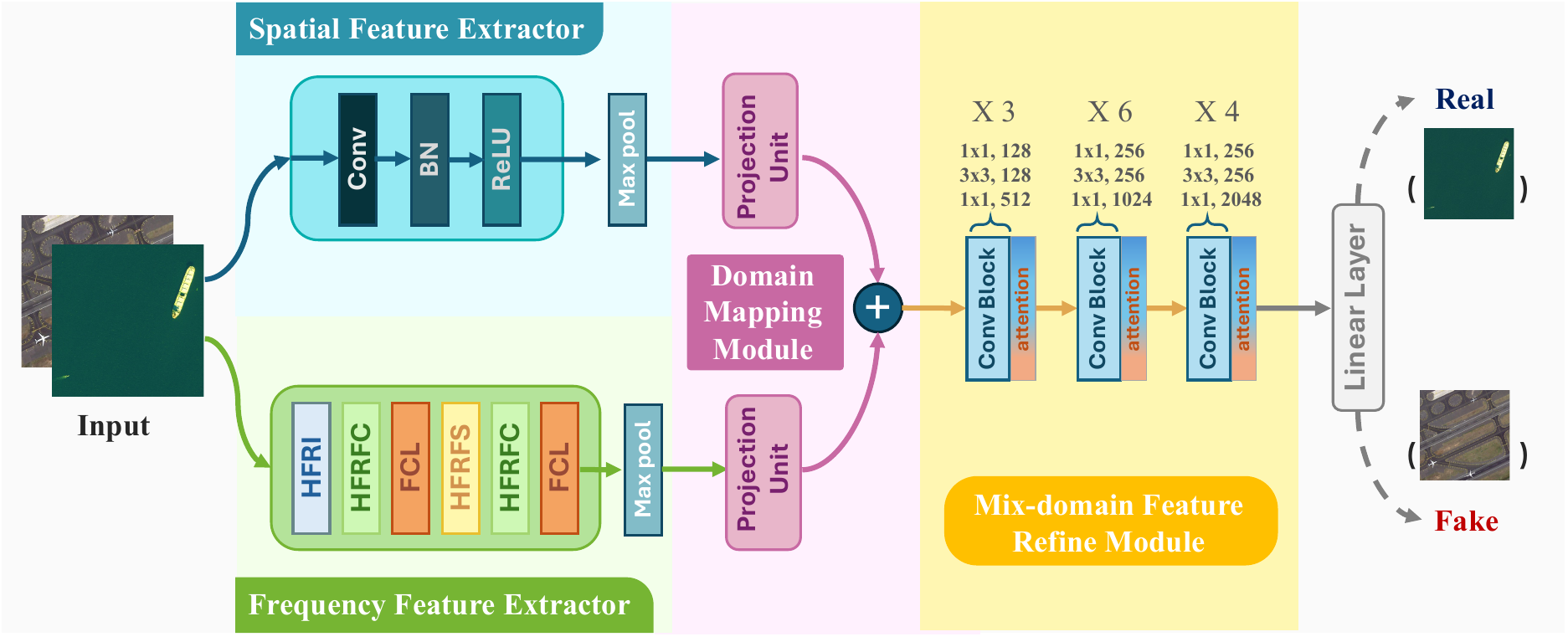}
  \caption{Overview of the proposed SFNet framework. For the input image, SFNet extracts rich features using spatial and frequency feature extractors. The two domain features are then mapped for initial alignment through the domain mapping module and fused by summing. Next, the mix-domain feature refinement module is employed to further fuse the two domain features and suppress redundant information.}
  \label{fig:overview}
\end{figure*}

First, the differentiation feature extraction branch is designed considering the difference between spatial domain and frequency domain. Specifically, we use a typical CNN structure that is good at feature extraction in the spatial domain, while for the frequency domain we transform the image into the phase spectrum and the amplitude spectrum before applying the convolutional layer. This design ensures that the network can fully exploit the artifacts representation.

Second, in order to avoid significant inconsistencies in feature fusion between image and frequency domains, we introduce a set of projected residual convolution groups that contain attention mechanisms for each extractors, and then fuse the features of the two domains. Afterwards, we use three groups of residual convolutions to deeply extract mixed artifact representation. Finally, the features are fed into a simple linear classifier to obtain a judgment about the authenticity of the image.

\subsection{Domain Encoder}

\subsubsection{Image Extractor}
In order to capture the forgery patterns in the spatial domain, we introduce a feature extraction layer to represent and learn the local feature coding of remote sensing images (as shown in Figure \ref{fig:overview}). In particular, we deliberately use simple single-layer convolution to construct the spatial domain extractor. This is because high-level semantic features are usually independent of local features since they are too abstract and shared among real and forged images, which is not conducive to characterizing artifact in images\cite{Liu_Li_Zhou_Chen_He_Xue_Zhang_Yu_2021}. Therefore, we design simple convolution to capture image-domain features and enhance the separability of authentic and fake images.

\begin{equation}
  f = MaxPool(ReLU(BN(conv(x))))
\end{equation}
Where $x\in \mathbb{R} ^{(C \times W \times H)}$ denotes input image, $Conv$ denotes convolution operation, $BN$ denotes Batch normal, $ReLU$ denotes activation, $Maxpool$ denotes maximum pooling operation.

\subsubsection{Frequency Extractor}

A large number of studies have shown that the high-frequency information in images is very important for the recognition of forged traces\cite{Qian_Yin_Sheng_Chen_Shao_2020},\cite{Tan_Zhao_Wei_Gu_Liu_Wei_2024},\cite{Jeong_Kim_Min_Joe_Gwon_Choi_2022}. However, in the original spatial domain, different frequency information is coupled with each other, resulting in the high-frequency component, which can provide key information for forged detection being covered up and difficult to be fully mined. Thus, we designed frequency extractor as shown in Figure \ref{frequency}. Firstly, frequency decoupling and low-frequency filtering are carried out at different levels to extract the high-frequency information of the image fully. Then we introduce a special frequency convolution module to learn the deep feature of the phase spectrum and amplitude spectrum of high-frequency information directly. Specifically, the frequency extractor consists of the following four submodules:

\begin{figure}[]
  \centering
  \includegraphics[width=0.99\linewidth]{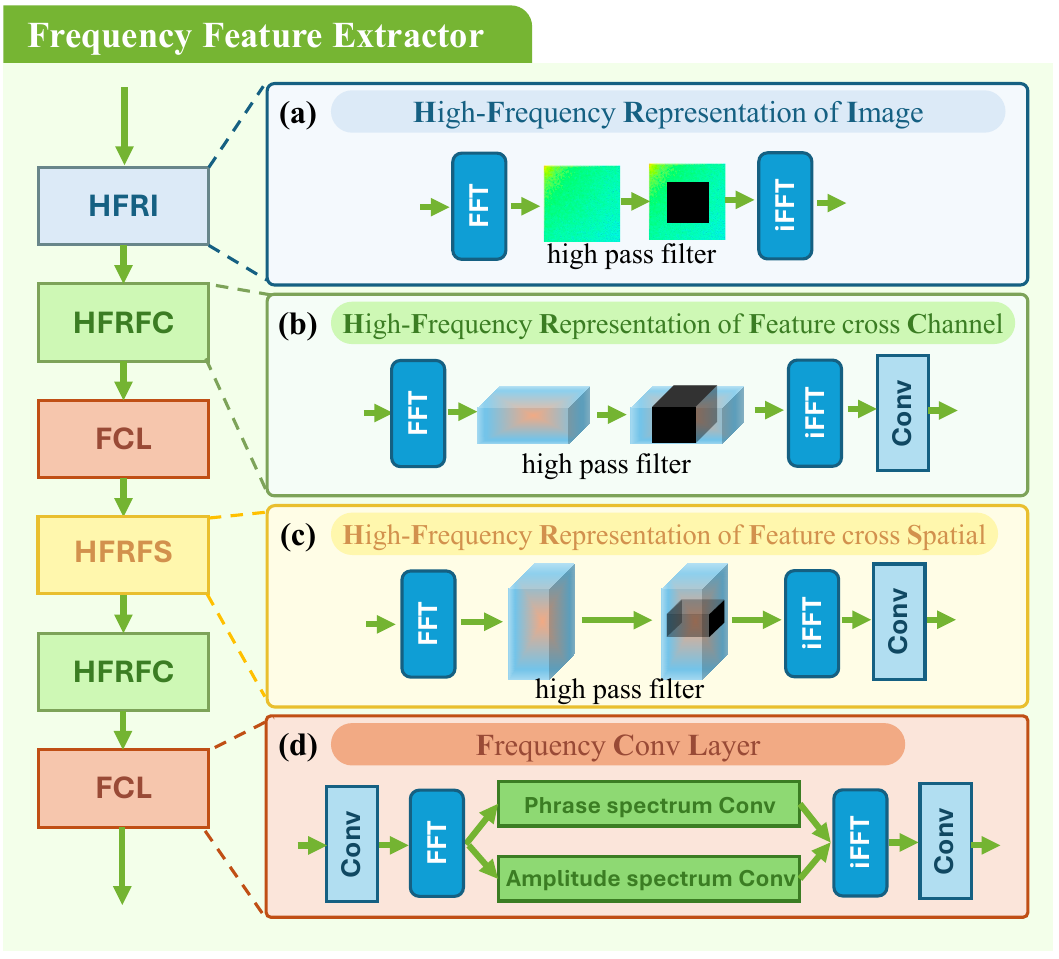}
  \caption{Frequency Extractor. For the input image, we first convert it to phase spectrum and the amplitude spectrum and then apply a convolutional layer so that the detector is able to learn in the frequency domain. The extractor contain four submodules as shown in the figure above.}
  \label{frequency}
\end{figure}

\paragraph{High-Frequency Representation of Image(HFRI)}
High-frequency artifacts are valuable indicators for distinguishing between real and forged images. Thus, we introduce the submodule HFRI to decompose the image into frequency domain frequency amplitude and phase information and retain high frequency information.(as shown in Figure \ref{frequency} a. For $x\in \mathbb{R}^{(C\times H\times W)}$ input image HFRI submodule operation process is as follows:

\begin{equation}
  x' = iFFT(B(FFT(x)))
\end{equation}

where $FFT()$ denotes the fast Fourier Transform (FFT) function to transform the image from the spatial domain to the frequency domain and move the zero-frequency component to the center. FFT of an RGB image results in the same vector dimension as the original image but with plural values, each representing amplitude and phase information for a particular frequency. $iFFT()$ denotes that the inverse fast Fourier transform (iFFT) converts the extracted high-frequency information back to the spatial domain. $B()$ denotes the high frequency information extracted from the representation transformed by applying a high pass filter (HPF), which is rich in image edge and texture details. The calculation rules are as follows:

\begin{equation}
  B(f_{i,j})=\left\{\begin{array}{cc}
  f_{i,j} & otherwise \\
      0 & if |i|<W_f/4,|j|<H_f/4
  \end{array}\right.
\end{equation}
where $x\in \mathbb{R}^{(3\times H\times W)}$represents the data processed by this submodule and describes the high-frequency components of this image. $W_f$ and $H_f$ denote the width and height of input $f_{i,j}$. This submodule's core significance is to highlight and represent the high-frequency information of the image, which is the key to distinguishing between real and forged images.

\paragraph{High-Frequency Representation of Feature cross Channel/Spatial}
Different generation approaches have different fake frequency patterns, while the frequency patterns of the single dataset are limited. Directly introducing frequency information to the model can easily cause overfitting, thus harming its generalization ability. To address this, we force the frequency encoder to prioritise high-frequency information in the feature space. Specifically, given any input feature $f$, HFRF-C (Figure \ref{frequency} (b)) and HFRF-S (Figure \ref{frequency} (c)) serve as two versions of HFRF that perform high-frequency enhancement in the channel and spatial (i.e., height and width (H,W))dimensions, respectively. The operations are performed as Eq.\ref{equ:equ_HRFCS}

\begin{figure*}[!b]
  \centering
  \begin{equation}
    \label{equ:equ_HRFCS}
    f'(dim)=\left\{\begin{array}{cc}
    Conv(iFFT_{Spatial}((B(FFT_{spatial}(f)))))& if with HFRF-S \\
        Conv(iFFT_{Channel}(B(FFT_{Channel}(f)))) & if with HFRF-C
    \end{array}\right.
  \end{equation}
  \label{hfrf}
  \end{figure*}

Among them, $FFT_{Spatial}$ and $iFFT_{Spatial}$ denote performing cross-space Fast Fourier Transform and inverse FFT transform. Similarly,  $FFT_{Channel}$ and  $iFFT_{Channel}$ represent the cross-channel FFT and its inverse. After the HFRF-S and HFRF-C submodules, the high-frequency information of the intermediate features can be further purified and strengthened to support the forged trace identification.

\paragraph{Frequency Conv Layer}
In the original image, the frequency information is often highly coupled, which makes it difficult for the traditional convolution layer to extract the high-frequency information and thus may overfit the insignificant low-frequency information in the data. To address this, we propose the FCL submodule to directly analyse the features of the high-frequency components of the image(as shown in Figure \ref{frequency} d). For input feature $y \in \mathbb{R} ^{C\times H \times W}$ , the operation process of FCL submodule is expressed as follows:
\begin{equation}
  \tilde{f} = FFT(Conv_1(f))
\end{equation}
\begin{equation}
  \hat{f}=Conv_{phase}(\tilde{f})+Conv_{amplitude}(\tilde{f})
\end{equation}
\begin{equation}
  f'=Conv_2(iFFT(\hat{f}))
\end{equation}

Specifically, the input feature $f$ is first transformed to the frequency domain by the fast Fourier transform (FFT). The resulting frequency domain feature $\hat{f}$ is a complex tensor, which can be further decomposed into phase and amplitude spectra. The former describes the spatial distribution of different frequency components, and the latter records the proportion of each frequency component in the image. Then, considering the fundamental physical differences between the two, we set two convolutional layers, $conv_{phase}$ and $conv_{amplitude}$, with the same structure but non-shared parameters, to be applied to the phase spectrum and the amplitude spectrum, respectively. Finally, the inverse Fast Fourier transform was used to convert the learned frequency information back to the standard feature space, to facilitate the subsequent fusion and complementation with spatial domain features.

\subsection{Cross-domain feature fusion}
To fully exploit the complementary effects of spatial domain and frequency domain features, we also proposed a Domain Mapping module and a Mix-domain Feature Refine module. The former aims to overcome the feature conflict caused by the notable differences between image and frequency domains by unifying multi-domain features into a common feature subspace, thus achieving a preliminary fusion of the two. The latter is a module that introduces an attention mechanism based on a standard residual convolutional network. On the one hand, it further optimizes the combination of low-level image and frequency features extracted by the encoder, and on the other hand, it helps the model learn to focus on the more critical features as clues adaptively.

\subsubsection{Domain Mapping Module}

The difference in content and physical meaning between spatial domain and frequency domain features is also the basis for their complementarity. If the two features are simply strictly aligned and fused, it will inevitably destroy their unique key information, and eventually lose the significance of fusion because of damage to complementarity. Therefore, how to ensure "seeking common while reserving differences" is the core issue to achieving effective multi-domain feature fusion. In this regard, the design process of the Domain Mapping module mainly follows the following two principles:

 \paragraph{Higher nonlinearity to improve the ability of feature transformation}: We set up three feature projection units in the multi-domain representation unified mapping module, where each unit is composed of a number of 3x3 and 1x1 convolutions stacked to achieve complex feature mapping. In addition, a nonlinear activation function ReLU is introduced between the convolutional layers to increase the overall nonlinear feature transformation ability.

 \paragraph{Provide shortcut connections to ensure key information is preserved}: In order to avoid the loss of key information in complex feature transformation, we introduced He et al. \cite{he2016deep} and set shortcut connections across projection units, so that some underlying features have the opportunity to be completely preserved.

\begin{figure}[]
  \centering
  \includegraphics[width=0.99\linewidth]{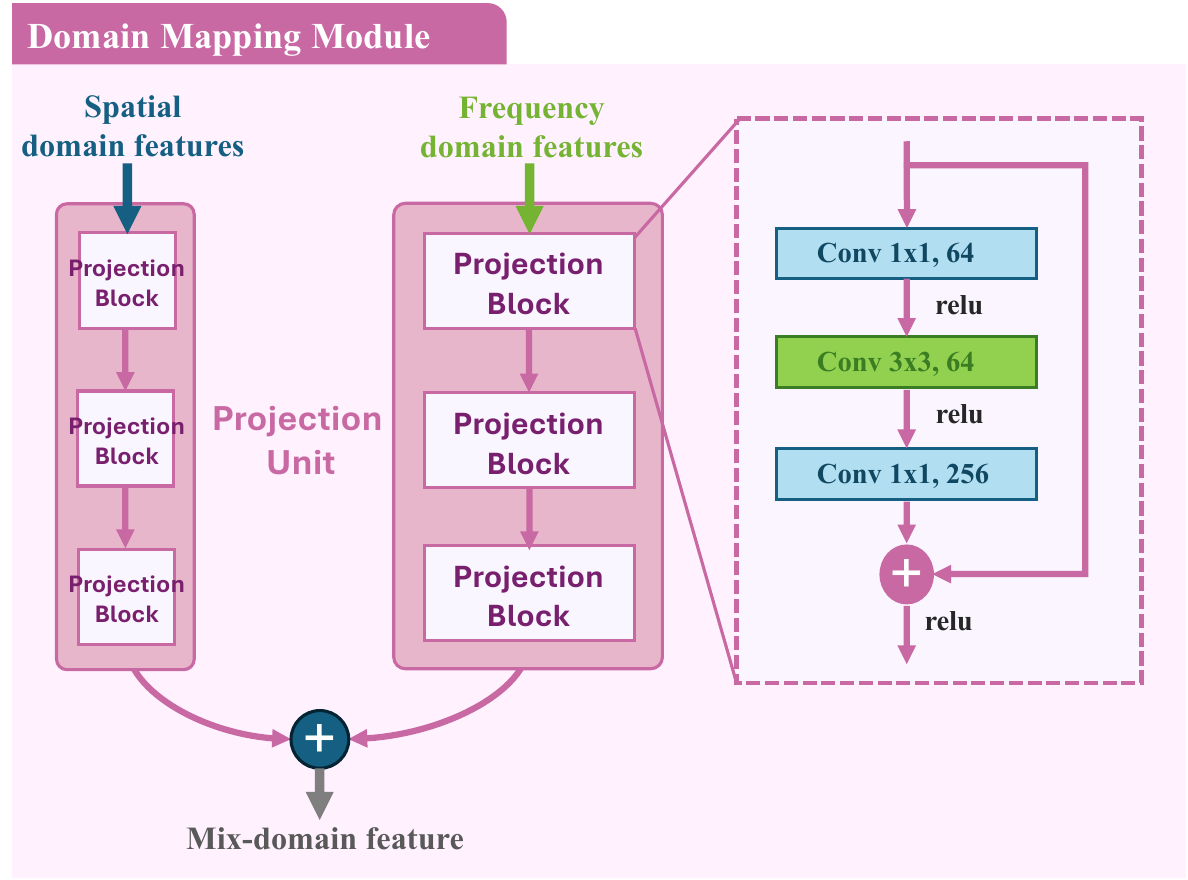}
  \caption{Domain Mapping Module. The Domain Mapping Module contains 2 projection units designed independently for the image and frequency domains. The projection unit structure is composed of three projection blocks, which includes the convolution structure with strong nonlinear representation ability and the skip structure that can ensure the integrity of information.}
  \label{domain}
\end{figure}

Based on the above two principles, we set up two projection unit with the same structure but non-shared parameters for spatial domain and frequency domain features(Figure \ref{fig:overview}), and the details of each branch are shown in Figure \ref{domain}.

\subsubsection{Mix-domain Feature Refine Module}

After the domain mapping module, the spatial domain and frequency domain features can be initially aligned added and fused, but there are still problems with how to make full use of their complementary roles. Considering the complexity of remote sensing images and the diversity and concealment of forged traces, the underlying features extracted by the encoder are rich but not always effective, and there may even be some contradictions between each other. In this regard, in addition to integrating and optimizing these basic features, another important purpose of the mix-domain refine module is to realize adaptive attention to the key features.

Therefore, we first construct the backbone part of the hybrid domain feature optimization model by multi-layer residual convolution(Figure \ref{fig:overview}). On this basis, we draw on the work of Woo et al.\cite{Woo_Park_Lee_Kweon_2018} to introduce a cross-spatial and cross-channel dual attention mechanism to selectively emphasize important features in different dimensions. Specifically, for the intermediate feature $f \in \mathbb{R} ^{W\times H \times C}$ output of the input convolutional layer, the overall calculation process of the attention mechanism is as follows:

\begin{equation}
  f'=M_{channel}(f)\otimes f
\end{equation}
\begin{equation}
  f'' = M_{spatial}(f')\otimes f'
\end{equation}

\paragraph{Channel Attention}: We exploit the inter-channel relationship by generating a channel attention map $M_channel(f)$. It describes which channel of the input feature $f$ we should focus on. Specifically, we first utilize the average pooling and max pooling operations to aggregate the channel information thus to generate two channel context descriptors: $F_{avg}^c$ and $F_{max}^c$. Then, these two descriptors will be forwarded to a shared network built by multilayer perceptron (MLP) to generate our channel attention map $M_{channel}(f)\in \mathbb{R}^(C\times1\times1)$. The whole calculation process of channel attention is shown as follow:

\begin{equation}
  \begin{aligned}
    M_{channel}(f) &=\sigma(MLP(AvgPool(f))+MLP(MaxPool(f))) \\
    &= \sigma(W_1(W_0(F^c_{avg}))+W_1(W_0(F^c_{max})))
  \end{aligned}
\end{equation}
where $\sigma$ denotes the sigma activate function. $W_0$ and $W_1$ denote learnable weight parameters of the MLP.

\paragraph{Spatial Attention}: We use the spatial relationship between different feature regions to generate the spatial attention map $M_{spatial}(f')$. Specifically, we summarize the channel information of the feature map by two pooling operations to generate two two-dimensional graphs: $F_{avg}^s\in \mathbb{R}^(1\times W\times H)$ and $F_{max}^s\in \mathbb{R}^(1\times W\times H)$. These features are then concatenated and convolved again using a standard convolutional layer to generate our 2D spatial attention map. The whole calculation process of spatial attention is shown as follow:

\begin{equation}
  \begin{aligned}
    M_{spatial}(f) &=\sigma(Conv^{7\times 7}(AvgPool(f));MaxPool(f))\\
    &= \sigma(Conv^{7 \times 7}([F_{avg}^s;F_{max}^s]))
  \end{aligned}
\end{equation}
where $\sigma$ denotes the sigma activate function. $Conv{7\times 7}$ denote the convolution operation with a filter size of ${7\times 7}$.

\section{Experiments}
This section consists of four main parts: 1. Experimental setup, where we introducted our experiment design; 2. Forgery Image Detection Experiment, where we compare SFNet with baseline models from the perspectives of detection accuracy and generalization ability; 3. Capture Visualization, where we visualize the false patterns captured by different models; 4. Exploring the Beneficial Features for RSI Forgery Detection, where we examine which features are most effective for detecting forgeries in remote sensing images.

\subsection{Experiments Setup}
\subsubsection{Datasets}
\begin{figure}[]
  \centering
  \includegraphics[width=0.9\linewidth]{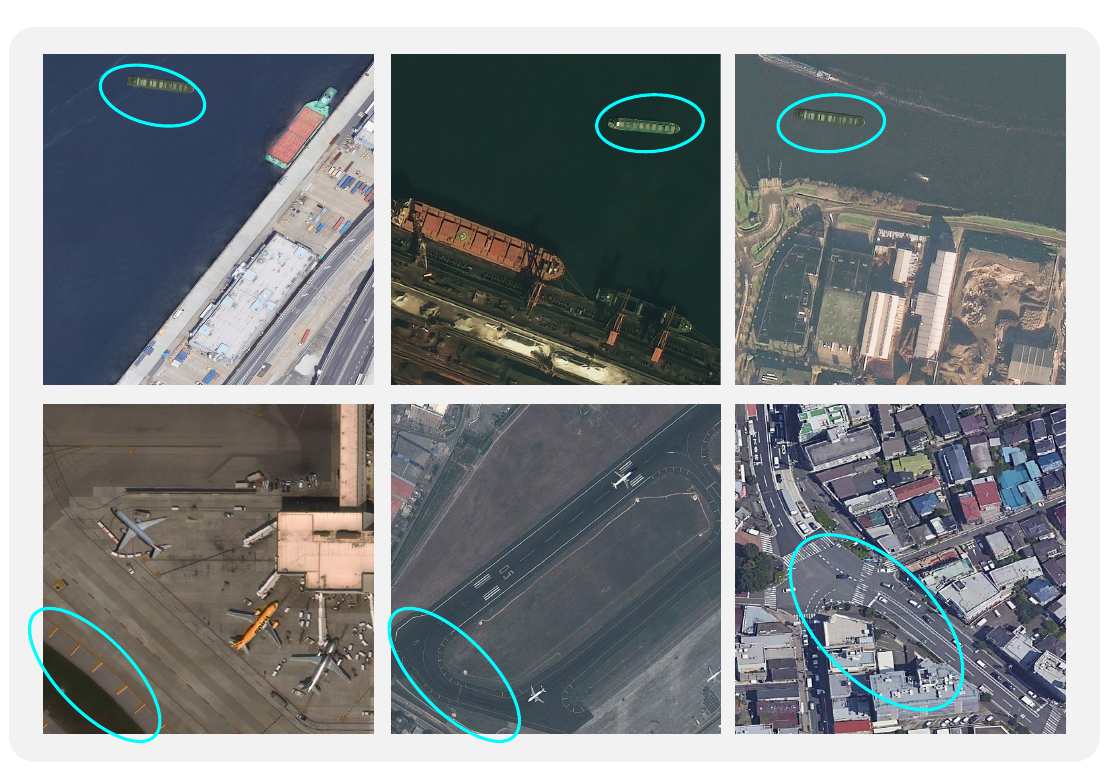}
  \caption{Forged image of ISPRS-FD, the blue circle represents the part of the artifacts in the false image.}
  \label{fig:isprs_fd}
\end{figure}

Three datasets were used in the experiment, as shown in the Table \ref{data}.
The basic dataset used in this study is ISPRS forgery detection (ISPRS-FD) \footnote{https://www.gaofen-challenge.com/challenge/dataset/14}. It has the following characteristics: 1. Diverse target scenarios: typical targets include ships, vehicles, etc., typical backgrounds include civil airports, sea, land, etc.; 2. Diverse resolution : the image size ranges from 256-20000 pixels, a total of 4742 images; 3. Various forging methods: including fake target editing, generation and modification of the diffusion model.
Figure \ref{fig:isprs_fd} presents some examples of fake images in ISPRS-FD that are supposed to be forged by different methods. The three images in the top of Figure \ref{fig:isprs_fd} cover different areas but contain identical ships, implying that they were faked by means of collage modification.
The three images in the bottom of Figure \ref{fig:isprs_fd}, although realistic, can be inferred from the unnatural distortions of the local lines that they should have been generated by advanced diffusion modelling.
\begin{table}[]
  \centering
  \caption{Datasets}
  \label{data}
  \begin{tabular}{c|c|c}
  \toprule
  \textbf{Dataset}                 & \textbf{Domain} & \textbf{Forgery Method}            \\
  \midrule
  ISPRS-FD                            & -               & Hybrid forgery                    \\ \hline
  \multirow{2}{*}{Anti DeepFake \cite{Zhao_Zhang_Xu_Sun_Deng_2021}}     & Beijing         & \multirow{2}{*}{Cycled-GAN}       \\
                                   & Seattle         &                                   \\ \hline
  \multirow{2}{*}{SDGen-Detection} & Potsdam         & \multirow{2}{*}{Stable-Diffusion} \\
                                   & Vaihingen       &     \\
                                   \bottomrule
  \end{tabular}
\end{table}

In addition, to measure the generalization ability of the model, we also used the Anti DeepFake dataset constructed by Zhao et al. \cite{Zhao_Zhang_Xu_Sun_Deng_2021} using the GAN technique and the SDGen-Detection dataset constructed by ourself using diffusion model. The details of the process of constructing SDGen-Detection dataset are shown in the Appendix \ref{sec:sdgen_con}.

Figure \ref{sdg} displays visualizations of images from the Anti DeepFake and SDGen-Diffusion dataset. The green circles highlight the impact of the long imaging distance on remote sensing images, which can easily introduce distortion during the preprocessing imaging process. Furthermore, the imaging mode used for RSI increases the likelihood of complex textures being produced.

\begin{figure*}[]
  \centering
  \includegraphics[width=0.9\linewidth]{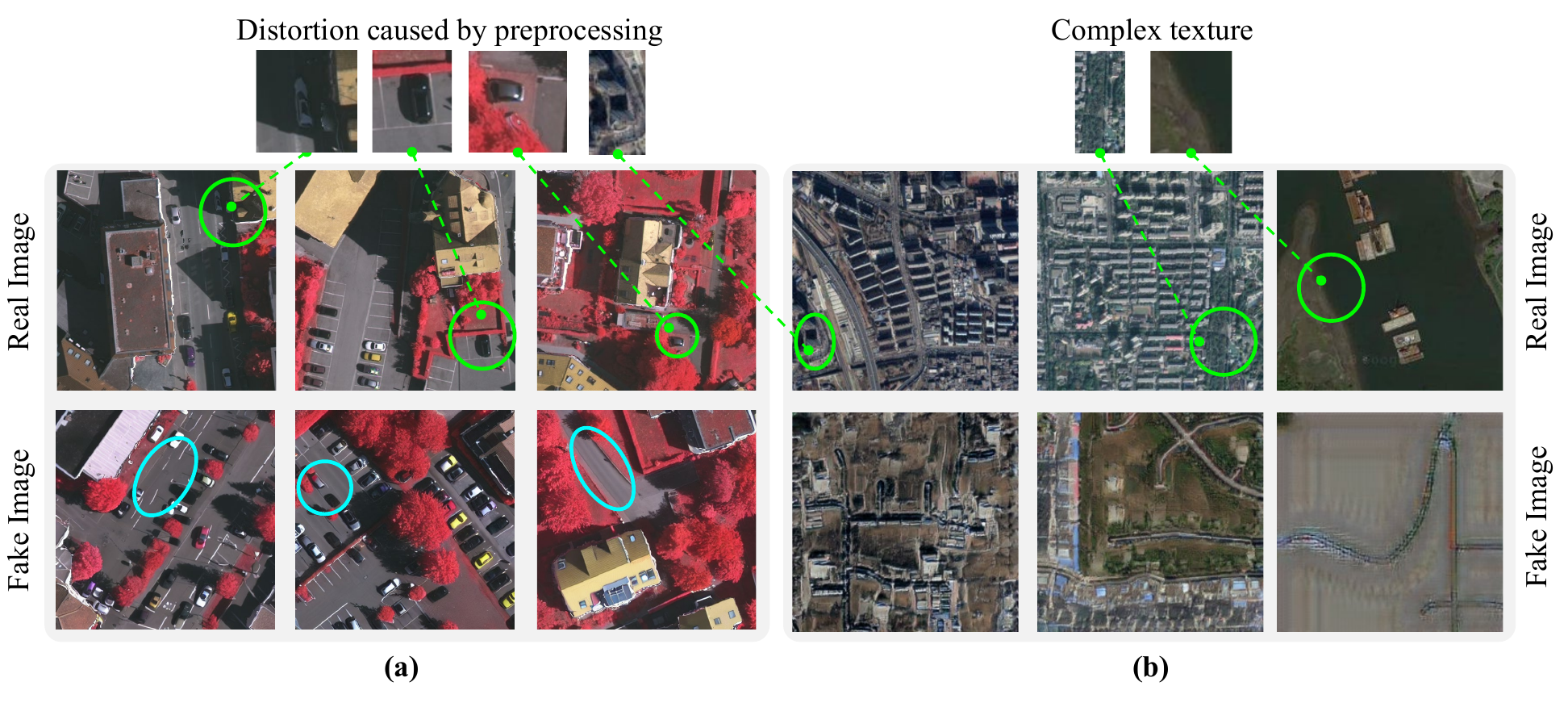}
  \caption{Real and fake image visualization. The green circle represents the part of the real image that can easily be mistaken for artifacts, and the blue circle represents the part of the artifacts in the false image. }
  \label{sdg}
\end{figure*}

\subsubsection{Baselines and metrics}
We collect the following information in binary problems: The number of samples predicted to be true by the true image (TP), the number of samples predicted to be true by the fake image (TP), the number of samples predicted to be fake by the true image (FN), the number of samples TN predicted by the fake image, using the research indicator over accuracy , Precision, Recall, f1-score comprehensively reflects the model’s ability to distinguish true and false images.

Our baseline approach falls into three main categories: Methods focused on the frequency domain, including Freqnet \cite{Tan_Zhao_Wei_Gu_Liu_Wei_2024}, FcaNet \cite{Qin_Zhang_Wu_Li_2021}, methods focused on the spatial domain, including Vision Transformer \cite{Dosovitskiy_2021}, Inception-v3 \cite{Szegedy_Vanhoucke_Ioffe_Shlens_Wojna_2016},  Conformer \cite{Peng_Huang_Gu_Xie_Wang_Jiao_Ye_2021}, and hybrid domain detection methods specialized in remote sensing, Geo-DefakeHop \cite{Chen_Zhang_Hu_You_Kuo_2021} and FSLNet \cite{Jia_Yao_2023}.

\subsubsection{Training Detail}
All experiments were conducted using MMmpretrain framework\footnote{\href{https://github.com/open-mmlab/mmpretrain}{https://github.com/open-mmlab/mmpretrain}}. All data sets were randomly divided according to the ratio of 0.5 for training set and 0.5 for validation set. The input image has a size of 256×256 after a specific data enhancement operation, we use the adam optimizer with a momentum of 0.9, a batch size of 16 per GPU at training time, and an initial learning rate of 0.0001. All models are trained using a cosine decay strategy in 300 cycles, and the training is implemented on a single Nvidia RTX 4090 GPU.

\begin{table}[]
  \centering
  \caption{Forgery detection comparison on ISPRS-FD dataset}
  \label{isprs}
  \scalebox{0.82}{
  \begin{tabular}{c|l|cccc}
  \toprule
  \textbf{Domain}               & \textbf{Method}          & \textbf{OA}                            & \textbf{Precision}                     & \textbf{Recall}                        & \textbf{f1-score}                      \\ \midrule
                                & Vit-base                & \cellcolor[HTML]{B5D57F}71.62          & \cellcolor[HTML]{A8D27F}62.08          & \cellcolor[HTML]{63BE7B}51.96          & \cellcolor[HTML]{63BE7B}47.10           \\
                                & Conformer        & \cellcolor[HTML]{FDC67D}77.69          & \cellcolor[HTML]{FA8D72}\textbf{76.22} & \cellcolor[HTML]{F2E783}64.34          & \cellcolor[HTML]{F8E983}65.95          \\
                                & Inception-v3             & \cellcolor[HTML]{D6DF81}73.56          & \cellcolor[HTML]{D1DE81}66.6           & \cellcolor[HTML]{D5DF81}61.89          & \cellcolor[HTML]{DFE182}62.82          \\
  \multirow{-4}{*}{Spatial}       & Resnet                   & \cellcolor[HTML]{E6E382}74.53          & \cellcolor[HTML]{FCEA83}71.26          & \cellcolor[HTML]{ACD37F}58.35          & \cellcolor[HTML]{B9D67F}58.01          \\ \hline
                                & Freqnet                  & \cellcolor[HTML]{CCDC81}73.01          & \cellcolor[HTML]{D9E081}67.38          & \cellcolor[HTML]{8CC97D}55.55          & \cellcolor[HTML]{97CD7E}53.79          \\
  \multirow{-2}{*}{Frequency}   & FcaNet                   & \cellcolor[HTML]{EDE582}74.95          & \cellcolor[HTML]{E6E382}68.82          & \cellcolor[HTML]{FAE983}65.05          & \cellcolor[HTML]{F9E983}66.14          \\ \hline
                                & Geo-DefakeHop          & \cellcolor[HTML]{63BE7B}66.64          & \cellcolor[HTML]{63BE7B}54.44          & \cellcolor[HTML]{70C17B}53.1           & \cellcolor[HTML]{8FCA7D}52.67          \\
                                & FSLNet                   & \cellcolor[HTML]{D3DE81}73.43          & \cellcolor[HTML]{E1E282}68.28          & \cellcolor[HTML]{FDB57A}69.90           & \cellcolor[HTML]{FED17F}68.87          \\
                                & \textbf{SFNet}           & \cellcolor[HTML]{F8696B}\textbf{81.82} & \cellcolor[HTML]{F8696B}\textbf{78.00}    & \cellcolor[HTML]{F8696B}\textbf{76.17} & \cellcolor[HTML]{F8696B}\textbf{76.97} \\
                                & SFNet w/o att            & \cellcolor[HTML]{FEC87E}77.60           & \cellcolor[HTML]{FED981}72.47          & \cellcolor[HTML]{FCA477}\textbf{71.31} & \cellcolor[HTML]{FCAB78}71.82          \\
                                & SFNet-Image w/o att      & \cellcolor[HTML]{FED380}77.10           & \cellcolor[HTML]{FFE683}71.81          & \cellcolor[HTML]{FCAE79}70.55          & \cellcolor[HTML]{FDB57A}71.10           \\
                                & SFNet-Image with att     & \cellcolor[HTML]{FCB27A}78.57          & \cellcolor[HTML]{FDB87B}74.08          & \cellcolor[HTML]{FCB079}70.34          & \cellcolor[HTML]{FCAD79}71.66          \\
                                & SFNet-Frequency w/o att  & \cellcolor[HTML]{FECE7F}77.35          & \cellcolor[HTML]{FDC37D}73.54          & \cellcolor[HTML]{FFE784}65.84          & \cellcolor[HTML]{FFE383}67.49          \\
  \multirow{-8}{*}{Spatial\&Freq} & SFNet-Frequency with att & \cellcolor[HTML]{FB9975}\textbf{79.71} & \cellcolor[HTML]{FA8D72}\textbf{76.22} & \cellcolor[HTML]{FCAB78}70.74          & \cellcolor[HTML]{FCA377}\textbf{72.47}\\
  \bottomrule
  \end{tabular}
}
\end{table}

\subsection{Forgery Image Detection Experiment}

\subsubsection{Comparison of Model Detection Capabilities}

We use each detection method to learn on the ISPRS-FD training data set(50\%), and measure in the test set(50\%). The results are shown in Table \ref{isprs}. \footnote{The statistical results are displayed using color levels, with red representing relatively high values and green representing relatively low values} \footnote{Bold indicates the first two digits of the maximum value}

It's interesting to note the significant gap between the overall accuracy (OA) and f1-score in both Image-based and Frequency-based methods. This suggests that the models tend to classify most samples as fake images, leading to a high precision rate, while also missing some correct samples, resulting in a low recall rate. Specifically, for the Vit-base method, the 24.52\% difference between OA and f1-score indicates a notable imbalance. On the other hand, the performance of the fusion domain method appears more balanced, indicating its robustness in datasets without falsification hypotheses.

In addition,the f1-score of SFNet being 24.18\% and 8.1\% higher than Geo-DefakeHop and FSLNet respectively demonstrates the effectiveness of SFNet's feature projection layer and attention mechanism. These components help in mitigating the loss of fusion information caused by semantic-level inconsistencies and differences in scales. As a result, SFNet performs better in completing the detection task compared to direct fusion methods.

The ablation experiments provide further insight into the points mentioned. It's observed that when SFNet uses the independent Spatial domain or Frequency domain for false image detection, there is an approximately 5\% loss in f1-score, and these methods exhibit a notable gap between OA and f1-score. However, when the attention mechanism is removed from SFNet, the model experiences around 5\% f1-score loss, which is not as pronounced in models using the Spatial domain or the Frequency domain independently. This highlights the critical role of the attention mechanism in facilitating information fusion between the two domains.

\subsubsection{Comparison of Model Generalization Ability}

To further evaluate the generalization ability of forgery detection model, we examine the performance of different model in a cross-regional data setting. We set up two situations, one is trained in the Anti DeepFake Beijing dataset and tested in the Anti DeepFake Seattle dataset; the other is trained in the SDGen-Detection Potsdam optical data set and tested on the SDGen-Detection Vaihingen dataset. The results are shown in Table \ref{gen}.

It's worth noting that some methods with a presupposition for false imaging showed a substantial difference results in this experiment. 
For instance, the basic assumption of Geo-DefakeHop is that GAN can reproduce the low-frequency response of a synthesized image well enough to produce a realistic image. Therefore, this method addresses the properties of GAN and focuses on distinguishing the difference between the high-frequency components of real and fake images. This is a method with strong a priori assumptions. This makes Geo-DefakeHop perform extremely well on the GAN-faked fake dataset Anti DeepFake, yet underperform in generalization on the more complex hybrid fake dataset SDGen- Detection. In comparison, SFNet demonstrated stable and high performance in both generalization tests, indicating its robustness in different forgery modes and scenarios.

Besides, we find that for GAN fake image detection, the frequency domain method has obvious advantages, while for stable-diffusion fake image detection, the attention-adding spatial domain method is more stable. To some extent, this reflects the generation characteristics of different fake images.

\begin{table}[]

  \centering
  \caption{Comparison of model generalization ability}
  \label{gen}
  \scalebox{0.88}{
  \begin{tabular}{c|l|l|ll}
  \toprule{\textbf{Dataset}}                                                                                                                & \textbf{Domain}               & \textbf{Method} & \textbf{OA}                            & \textbf{f1-score}                    \\ \midrule
                                                                                                                                  &                               &  Vit-base        & \cellcolor[HTML]{63BE7B}64.19          & \cellcolor[HTML]{63BE7B}61.67        \\
                                                                                                                                  &                               &  Conformer  & \cellcolor[HTML]{FFEB84}83.65          & \cellcolor[HTML]{FFEB84}82.55        \\
                                                                                                                                  &                               & Inception-v3    & \cellcolor[HTML]{F0E683}81.85          & \cellcolor[HTML]{F1E683}80.69        \\
                                                                                                                                  & \multirow{-4}{*}{Spatial}       & Resnet          & \cellcolor[HTML]{E1E282}79.93          & \cellcolor[HTML]{E5E382}79.15        \\\cline{2-5}
                                                                                                                                  &                               & Freqnet         & \cellcolor[HTML]{F9776E}\textbf{98.35} & \cellcolor[HTML]{F9776E}98.25        \\
                                                                                                                                  & \multirow{-2}{*}{Frequency}   & FcaNet          & \cellcolor[HTML]{FFE082}85.1           & \cellcolor[HTML]{FFE483}83.56        \\ \cline{2-5}
                                                                                                                                  &                               & Geo-DefakeHop   & \cellcolor[HTML]{F8696B}\textbf{100}   & \cellcolor[HTML]{F8696B}\textbf{100} \\
                                                                                                                                  &                               & FSLNet          & \cellcolor[HTML]{CFDD81}77.67          & \cellcolor[HTML]{D8DF81}77.36        \\
  \multirow{-9}{*}{\begin{tabular}[c]{@{}c@{}}Anti DeepFake \\ (Training on Beijing\\  test on Seattle)\end{tabular}}                & \multirow{-3}{*}{Spatial\&Freq} & \textbf{SFNet}  & \cellcolor[HTML]{FB9774}94.26          & \cellcolor[HTML]{FB9674}94.01        \\\midrule
                                                                                                                                  &                               &  Vit-base         & \cellcolor[HTML]{FB9874}77.41          & \cellcolor[HTML]{FB9273}77.18        \\
                                                                                                                                  &                               &  Conformer  & \cellcolor[HTML]{FB9173}77.98          & \cellcolor[HTML]{FFE884}68.33        \\
                                                                                                                                  &                               & Inception-v3    & \cellcolor[HTML]{8DCA7D}65.95          & \cellcolor[HTML]{C3D980}65.80         \\
                                                                                                                                  & \multirow{-4}{*}{Spatial}       & Resnet          & \cellcolor[HTML]{98CD7E}66.35          & \cellcolor[HTML]{D1DD81}66.31        \\\cline{2-5}
                                                                                                                                  &                               & Freqnet         & \cellcolor[HTML]{F9726D}80.78          & \cellcolor[HTML]{F96F6D}80.76        \\
                                                                                                                                  & \multirow{-2}{*}{Frequency}   & FcaNet          & \cellcolor[HTML]{C5DA80}67.84          & \cellcolor[HTML]{E6E382}67.07        \\\cline{2-5}
                                                                                                                                  &                               & Geo-DefakeHop   & \cellcolor[HTML]{FFEB84}69.79          & \cellcolor[HTML]{FFEB84}68.00           \\
                                                                                                                                  &                               & FSLNet          & \cellcolor[HTML]{63BE7B}64.53          & \cellcolor[HTML]{63BE7B}62.19        \\
  \multirow{-9}{*}{\begin{tabular}[c]{@{}c@{}}SDGen-Detection  \\ (Training on Potsdam \\ and testing on Vaihingen)\end{tabular}} & \multirow{-3}{*}{Spatial\&Freq} & \textbf{SFNet }          & \cellcolor[HTML]{F8696B}81.59          & \cellcolor[HTML]{F8696B}81.32       \\
  \bottomrule
  \end{tabular}
}
\end{table}

\subsection{Artifacts Capture Visualization}

\begin{figure*}[]
  \centering
  \includegraphics[width=0.9\linewidth]{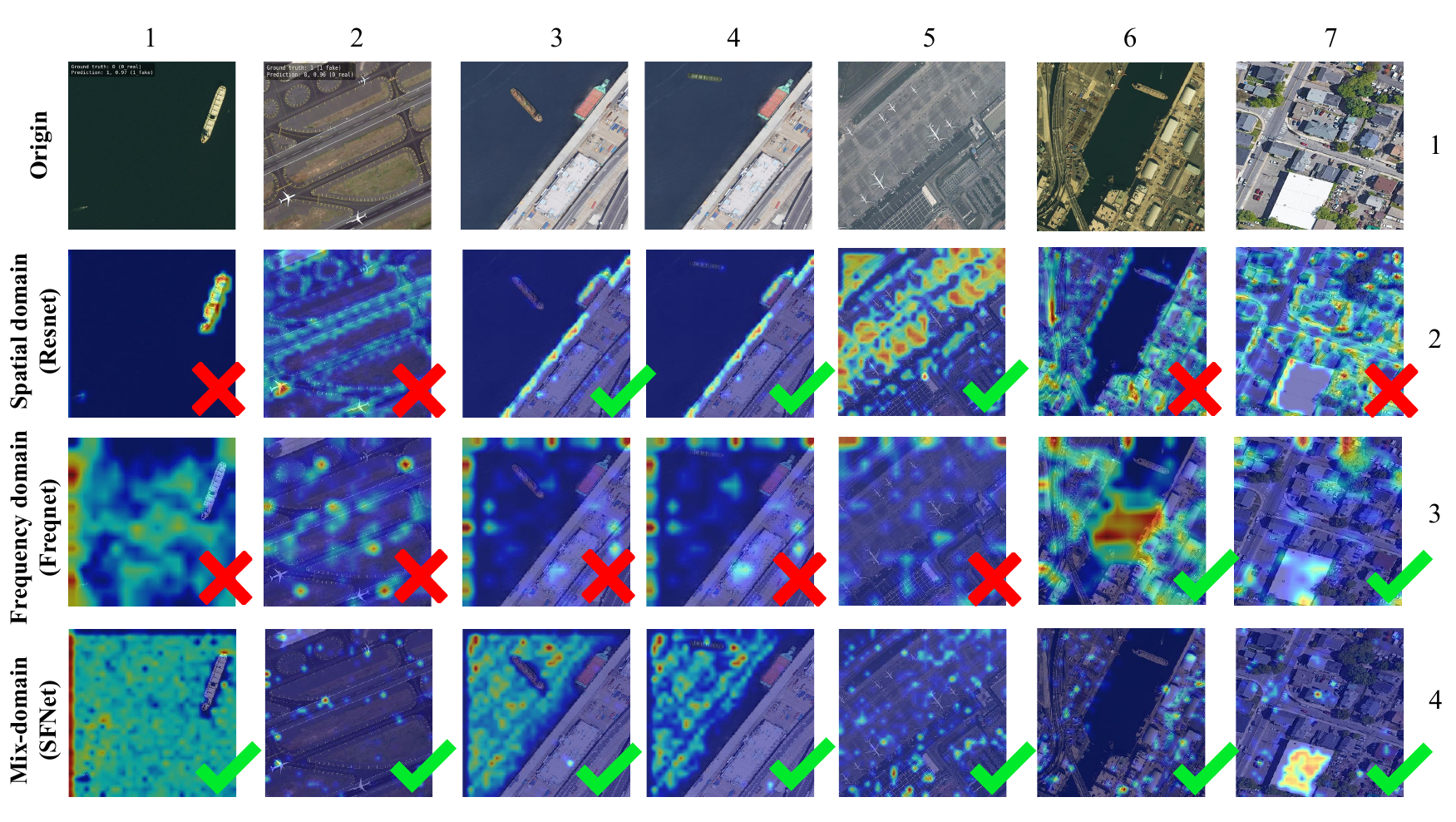}
  \caption{Visualizations of different domain detection models by GradCAM. All the remote sensing images in the figure are fake. A checkmark signifies that the model correctly identified the image as fake, while a cross denotes that the model incorrectly classified it as real}
  \label{cam}
\end{figure*}

In this section, we select the ISPRS-FD test data set for GradCAM visualization \cite{Selvaraju_Cogswell_Das_Vedantam_Parikh_Batra_2020} to further explore the false patterns captured by the spatial domain model(represented by ResNet), the frequency domain model(represented by FreqNet) and the mix-domain model(represented by SFNet) of model (see Fig \ref{cam}).
It is worth noting that the frequency domain model performs singly in false pattern detection, such as in Fig \ref{cam} row 3, column 3 through row 3, column 5((3, 3)-(3, 5)), whose edge responses are strong and similar. This indicates that the frequency domain has limitations in capturing a comprehensive range of forgery pattern. Furthermore, the spatial domain model exhibits restricted performance when dealing with complex scenes containing numerous objects and intricate information. For example,  in Fig \ref{cam} row 2, column 6 and row 2, column 7 ((2, 6), (2, 7)), the background information is complex, and the spatial-based detector cannot fully caputure the anomalies by only compressing the information. SFNet performs well in these scenarios. Dual-domain information capture ensures the ability to capture false patterns, and the attention mechanism ensures filtering focus, for example,  in Fig \ref{cam} row 4, column 7 and row 4, column 4((4, 7) (4, 4)).

\subsection{Exploring the Beneficial Features for RSI Forgery Detection}
We hope to further explore the nature of forgery patterns, i.e., for what type of information (semantic or image itself) and to what extent these patterns are beneficial. This is significant for both detection model optimization and design. Therefore, we tested the SFNet on the ISPRS-FD dataset with different pre-training weights (representing different information, specifically, self-supervised pretraining emphasizes the features of the image itself, and supervised pretraining emphasizes high-level semantic features), including:

\begin{itemize}
  \item Pre-training on natural image dataset:
  \begin{itemize}
    \item Pre-training weights from supervised learning on ImageNet \cite{Deng_Dong_Socher_Li_Li_Fei-Fei_2009}, a multi-million image classification dataset.
    \item Self-supervised comparative learning \cite{Chen_Kornblith_Norouzi_Hinton_2020} on ImageNet.
  \end{itemize}
  \item Pre-training on RSI dataset:
  \begin{itemize}
    \item Models from supervised pre-training on Million-AID \cite{Wang_Zhang_Du_Xia_Tao_2022} containing 100M remote sensing image scene classification samples.
    \item Models from self-supervised comparative learning on Million-AID
    \item Models pre-trained on TOV-RS \cite{Tao_Qi_Zhang_Zhu_Lu_Li_2023} with 100M remote imagery samples
    \item Self-supervised model for comparison on TOV-RS
  \end{itemize}
\end{itemize}

The results are shown in Table \ref{pretrain}.

\begin{table*}[]
  \centering
  \caption{Detection results of SFNet under different pre-training methods}
  \label{pretrain}
  \begin{tabular}{c|c|c|cccc}
  \toprule
  \textbf{Info-type}                       & \textbf{Domain-type} & \textbf{pretrain}    & \textbf{OA }            & \textbf{Precision}      & \textbf{Recall }        & \textbf{f1-score}             \\ \midrule
  -                               & -           & W/o pretrain & 81.82          & 78.00             & 76.17          & 76.97          \\ \hline
  \multirow{2}{*}{Spatial}          & SL          & \multirow{2}{*}{ImageNet}      & \textbf{82.58} & \textbf{78.94} & 77.28          & 78.02          \\
                                  & SSL         &     & 82.29          & 78.21          & \textbf{78.90}  & \textbf{78.54} \\ \hline
  \multirow{2}{*}{Remote Sensing} & SL          & Million-AID  & 80.35          & 75.95          & 75.09          & 75.49          \\
                                  & SSL         & TOV-RS       & \textbf{83.17} & \textbf{79.78} & \textbf{77.83} & \textbf{78.68} \\ \bottomrule
  \end{tabular}
  \end{table*}

  We noticed that unsupervised pre-training models are usually better than supervised pre-training models. For example, Million-AID an TOV-RS both trained on RSIs. While the supervised Million-AID pre-trained model is nearly 3\% lower than the self-supervised TOV-RS pre-trained model, which shows that in the RSI forgery detection, the information of the image itself is more important than the semantic information.

  In addition, the TOV-RS pretrain-model based on RSI achieve higher performance than Image-Net-baed model. Large categories pre-training model (Image-Net SSL 1000 class) is better than few categories pre-training model (Milli-AID 40 class). These indicates that rich semantic information is helpful to distinguish the truth from the falsity of the model, which proves the necessity of introducing Image-domain into SFNet.

  At the same time, we note that using only a few categories of remote sensing pre-trained models even has a negative impact on the detection ability of 1.47\%, which indicates that focusing only on semantic information will ignore forgery details, resulting in the limitation of model discrimination.

\section{Discussion}
This subsection focuses on the nature and false image characteristics of SFNet and contains four main part: 1. DCT visualization, where we visualize the characteristics of different forgery RSI dataset in frequency perspective; 2. Effect of data augmentation on the model, where we discuss the impact of data augmentation; 3. Filtering low-frequency information ablation experiment, where we tested the effect of removing different proportions of low-frequency information on the SFNet; 4. Mix-Domain refine module ablation experiment, We tested the effect of different attentional mechanisms.

\subsection{DCT Visualization}
\begin{figure*}[]
  \centering
  \includegraphics[width=0.8\linewidth]{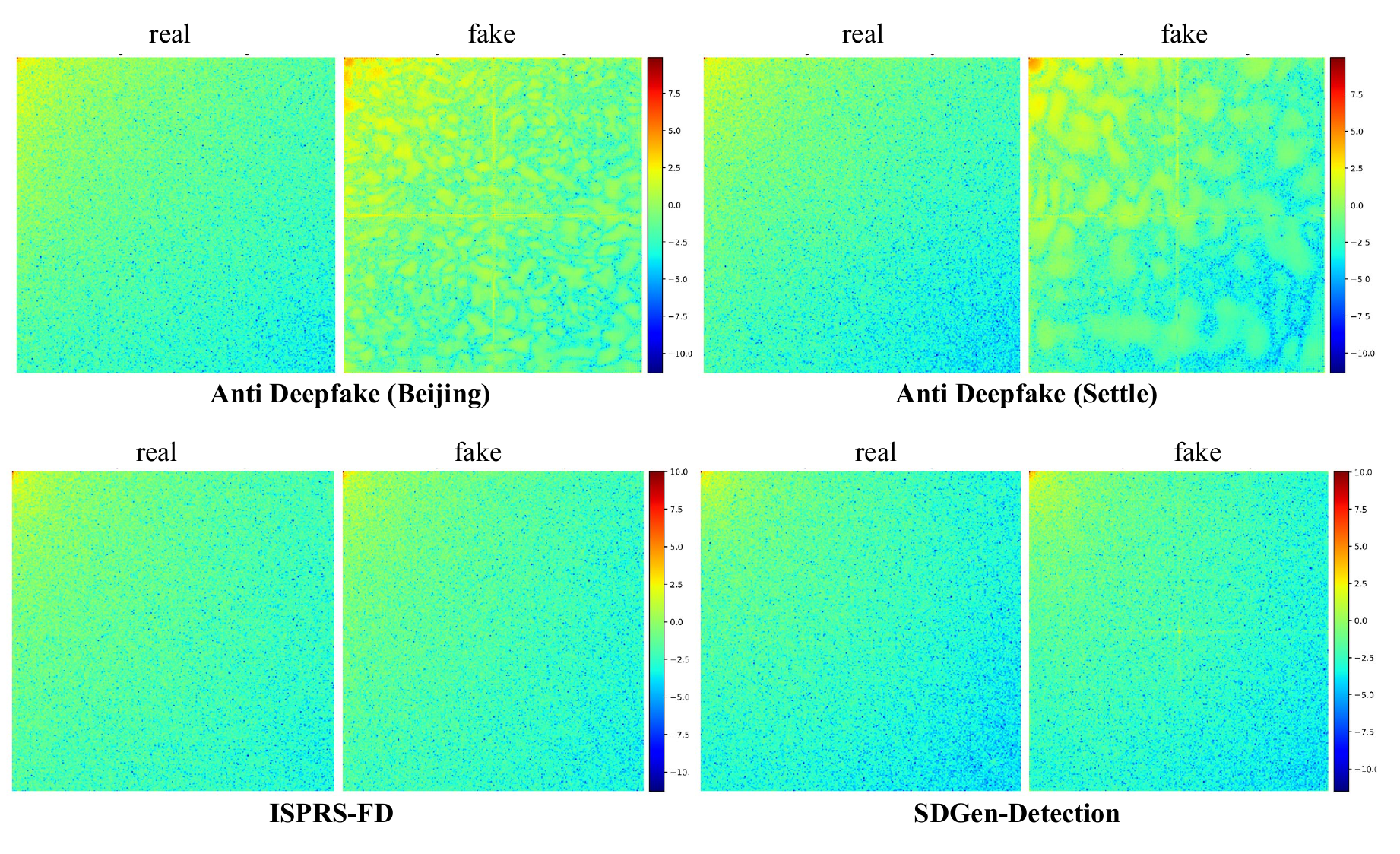}
  \caption{DCT visualization on datasets}
  \label{dct}
\end{figure*}

Figure \ref{dct} shows the frequency domain DCT visualization on RSI datasets. The frequncy feature of real and fake images on cycle-GAN-based datasets (Zhao (2021)) are large and easy to distinguish. ISPRS makes no assumptions about the method of forgery and is the most difficult to distinguish. The SG-Diffusion distinction is between the two data sets above.

\subsection{Effect of Data Augmentation on the Model}

\begin{figure}[]
  \centering
    \includegraphics[width=0.9\linewidth]{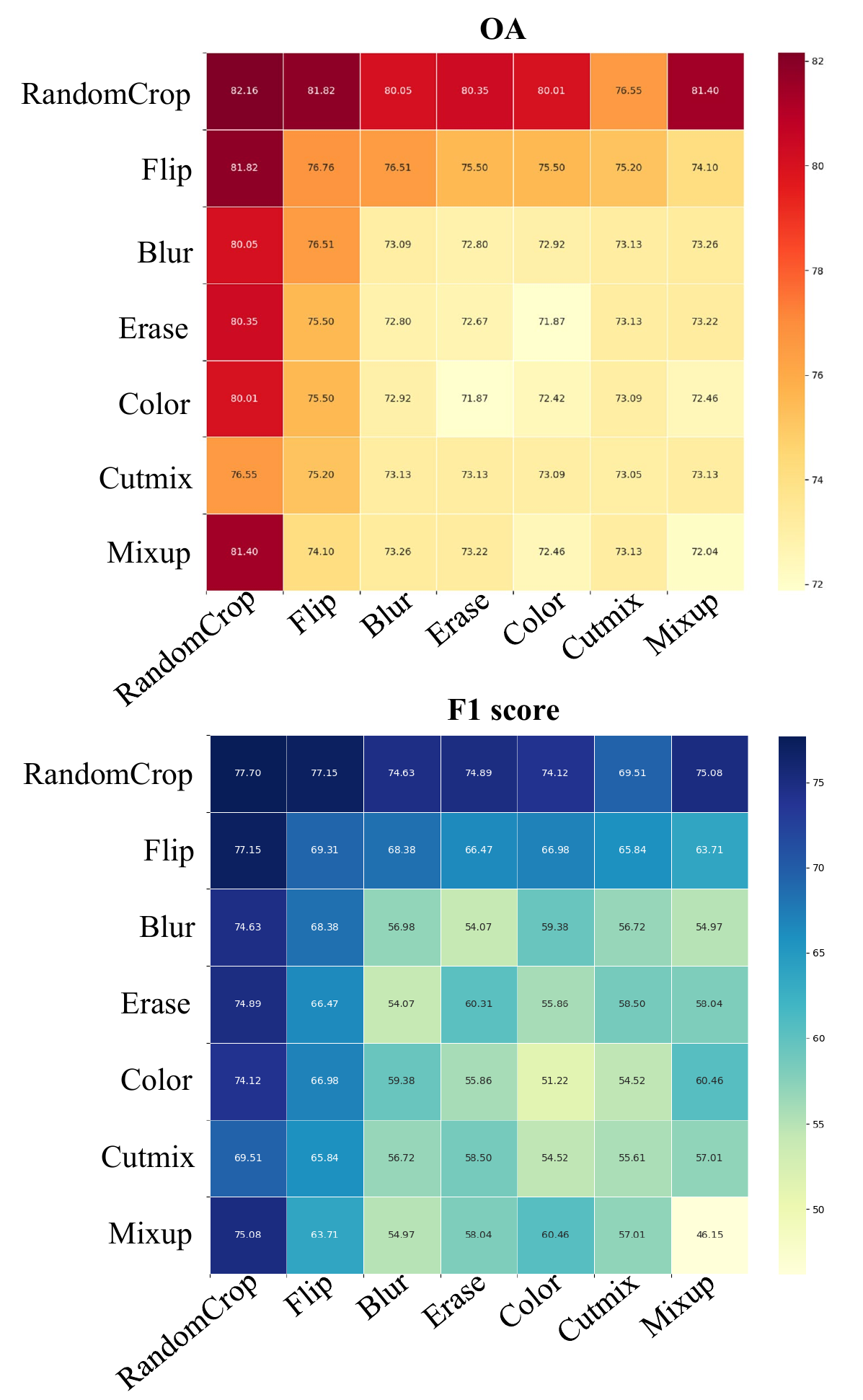}
    \caption{Comparison of different data Augmentation.}
    \label{aug}
  \end{figure}

Data augmentation is commonly used to increase the diversity of data in the spatial domain, but its direct application to the forgery detection can lead to the loss of crucial feature components and the introduction of outliers. This mismatched augmentation operation may cause the model to capture unreal features during learning, resulting in problems with truth-false recognition task confusion and performance degradation. We tested SFNet's detection capabilities under different data augmentation modes using the ISPRS dataset, as shown in Figure \ref{aug}.

Among them, the \textit{Random Crop} augmentation significantly improves the model's detection ability. On the other hand, other methods such as \textit{color}, \textit{inversion} and \textit{erasing} to some extent reduce the model's detection ability because they introduce noise that affects the model's ability to identify forgery patterns. Additionally, generation and fitting type augmentation methods, such as \textit{mixup}, which continuously fit discrete sample points to fit the real sample distribution, have the greatest impact on the model's detection ability, causing a 31.55\% decrease in the F1 score.

\subsection{Filtering Low-frequency Information Ablation Experiments}
\begin{figure}[]
  \centering
    \includegraphics[width=0.91\linewidth]{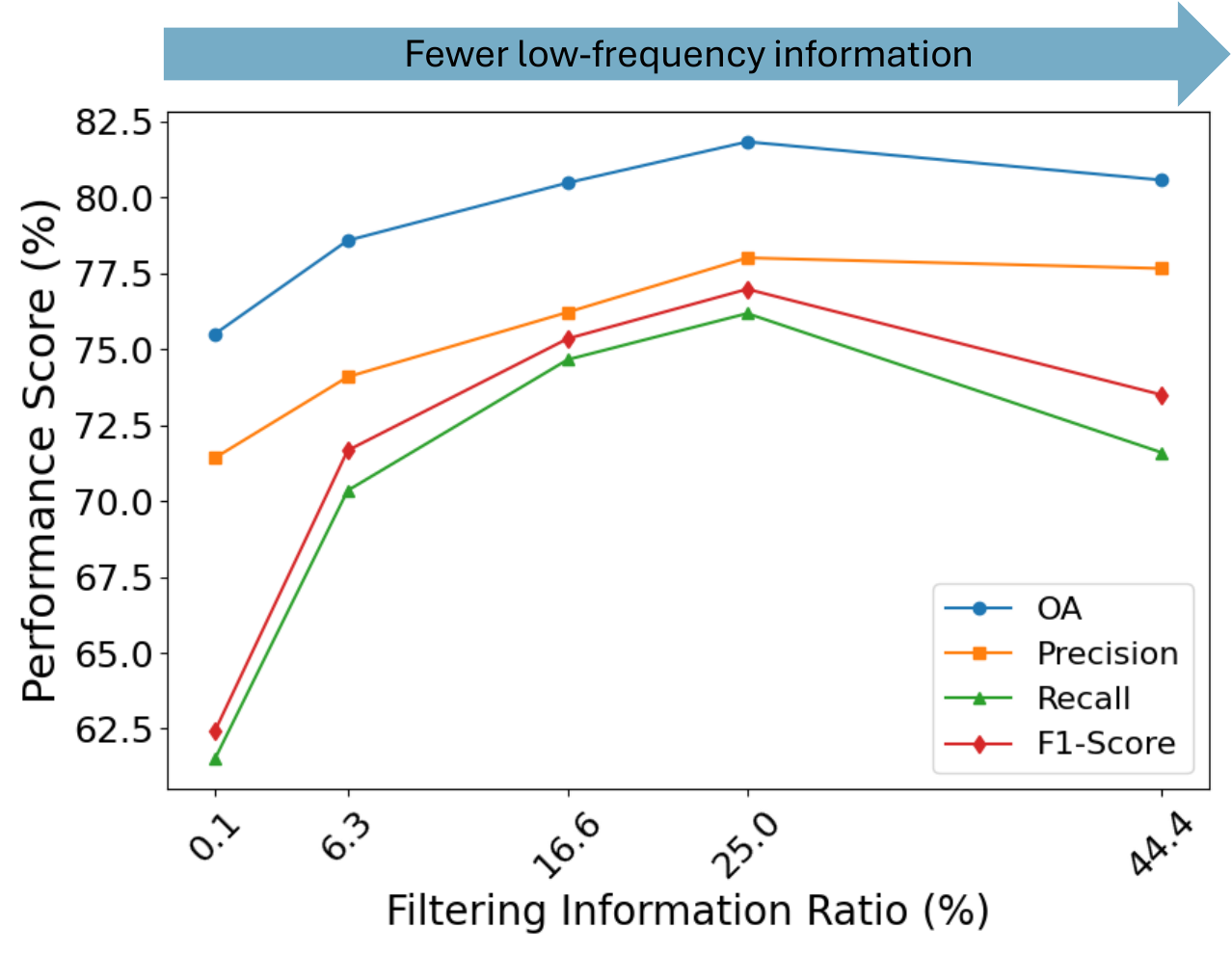}
    \caption{Filtering Low-frequency Information Ablation Experiments.}
    \label{scale}
\end{figure}

To verify the effectiveness of low-frequency information filtering, we conducted this experiment. In the frequency domain module of SFNet, information is arranged from high to low frequency, allowing us to apply a low-frequency filter. To evaluate the impact of low-frequency information, we control the ratio of low-frequency filtering by controlling the spectral cropping area in the high-frequency representation module. Specifically, the cropping area is represented as follows:
\begin{equation}
  [h/2-h/scale,h/2+h/scale]
\end{equation}
where h denotes the image hight. The relationship between scale and the percentage of filtered information is given in the following equation:
\begin{equation}
  \phi = 4/scale^2, \phi \in (0,1)
\end{equation}
where $\phi$ denotes the information filtering ratio.
When the scale is set to 2, the module is an extremely high-pass filter that loses almost all image structure. We set information filtering ratio to: 44.4\%, 25.0\%, 16.6\%, 6.3\%, and 0.1\% by setting the parameter scale to 3, 4, 5, 8, and 64, respectively. Figure \ref{scale} illustrates the model’s detection performance. The results show that when the information filtering ratio exceeds 25\% (scale$<$4), the detection performance decreases as the scale increases, indicating that an excess of low-frequency information negatively affects the model’s effectiveness. Conversely, when the information filtering ratio is higher than 25\% (scale$>$4), the detection performance improves with an increasing scale, suggesting that low- and mid-frequency information positively contributes to detection performance.

\subsection{Mix-Domain Refine Module Ablation Experiment}
\begin{table}[]
  \centering
  \caption{Mix-Domain Refine Module Ablation Experiment}
  \begin{tabular}{c|cccc}
  \toprule
  \multicolumn{1}{l|}{\textbf{}} & \textbf{OA}    & \textbf{Precision} & \textbf{Recall} & \textbf{f1-score} \\ \midrule
  CBAM(ours)                           & \textbf{81.82} & \textbf{78.00}        & \textbf{76.17}  & \textbf{76.97}    \\
  BAM                            & 79.88          & 75.80               & 72.41           & 73.69             \\
  SE                             & 79.92          & 75.54              & 73.69           & 74.48             \\
  ECA                            & 79.54          & 74.95              & 73.95           & 74.41             \\ \bottomrule  
\end{tabular}
\label{attention}
  \end{table}

In Table \ref{attention}, we present the results of an ablation study evaluating the impact of different attention mechanisms in the Mix-Domain Refine Module, specifically comparing CBAM\cite{Woo_Park_Lee_Kweon_2018}, BAM\cite{park2018bam}, SE\cite{hu2018squeeze}, and ECA\cite{wang2020eca}.  Among these, the Convolutional Block Attention Module (CBAM) demonstrates the best overall performance, achieving an overall accuracy (OA) of 81.82\%.  In comparison, the Bottleneck Attention Module (BAM) performs lower across all metrics, with an OA of 79.88\%.  The Squeeze-and-Excitation (SE) and Efficient Channel Attention (ECA) modules exhibit similar performance, with SE attaining an OA of 79.92\%, while ECA achieves an OA of 79.54\%.  These results highlight that CBAM significantly enhances the model’s performance, making it the most effective refinement module for this task.

\section{Conclusion}
In this paper, we presented SFNet, a novel approach for remote sensing image forgery detection that leverages the fusion of image and frequency-domain features. Our experimental results demonstrate that SFNet significantly outperforms existing state-of-the-art methods in terms of accuracy and robustness. By incorporating both spatial and frequency information, SFNet effectively captures the subtle inconsistencies introduced during the forgery process, leading to improved detection performance.
We evaluated the performance of SFNet on several benchmark datasets, where it consistently achieved higher detection rates compared to traditional methods. Additionally, our ablation studies confirmed the importance of each component within the SFNet architecture, highlighting the complementary nature of spatial and frequency-domain features in detecting forged regions. The proposed method's robustness to various types of forgeries and noise further emphasizes its potential applicability in real-world scenarios.
In brief, SFNet represents a significant advancement in the field of image forgery detection, offering a powerful and efficient tool for safeguarding the integrity of remote sensing imagery. We believe that the insights gained from this research will contribute to the development of more robust and versatile image analysis systems in the future.

%
\IEEEpeerreviewmaketitle

\appendix[Construction of SDGen-Detection dataset]
\label{sec:sdgen_con}

\begin{figure*}[]
  \centering
  \includegraphics[width=0.8\linewidth]{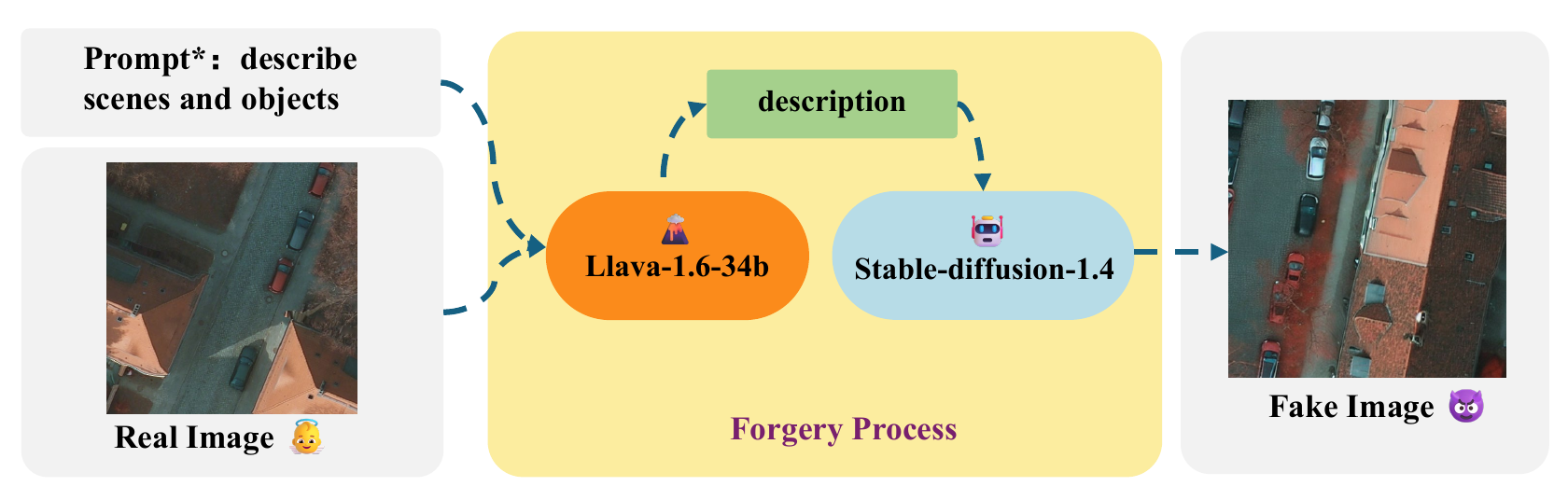}
  \caption{Fake image generation flow of sdgen-detection}
  \label{sdgen}
\end{figure*}

SDGen-Detection is generated based on the ISPRS Potsdam and Vaihingen datasets \cite{Rottensteiner_Sohn_Jung_Gerke_Baillard_Benitez_Breitkopf_2012}. The former is a typical historical city with large buildings, narrow streets and dense settlement structures; The latter is a relatively small village with many individual buildings and small multi-storey buildings. In particular, Vaihingen is not an ordinary optical image, but consists of three bands: near-infrared, red and green.
Our generation process is shown in Figure \ref{sdgen}. First, we input the real image and prompt (as shown in Table \ref{prompt}) into the Llava-1.6-34b visual-language model \cite{Liu_Li_Wu_Lee_2023,Lu_Li_Fu_Eckstein_Wang_2024} to generate the description of the image. Then, we input Stable-diffusion-1.4 for the description of the image to generate a fake image.

\begin{table}[]
  \caption{Prompt of Llava model for SDGen-Detection dataset}
  \label{prompt}
  \begin{tabular}{p{8.2cm}}
  \toprule
  \textbf{question}=
  You are an excellent visual language assistant who is able to describe scenes and objects ('impervious surface', 'building', 'low vegetation', 'tree', 'car') based on the content of an image (This is a near-infrared image with vegetation generally appearing red) with the following requirements:

  - At the \{pos\} of this image, describe the scene starting with: ``\{pos\} of image: ''

  - Your description can’t be ambiguous;

  - Do not repeat the answer to your description;

  - Your answer can’t be more than 20 words. \\

  \textbf{pos} ={[}'center','upper left','upper right','lower left', 'lower right'{]}\\
  \bottomrule
  \end{tabular}
\end{table}

%



\ifCLASSOPTIONcaptionsoff
  \newpage
\fi



\bibliographystyle{IEEEtran}
\bibliography{IEEEabrv,References.bib}
%

\begin{IEEEbiography}
  [{\includegraphics[width=1in,height=1.25in,clip,keepaspectratio]{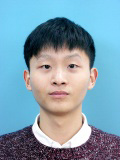}}]{Ji Qi} received a B.S. degree in remote sensing science and technology from Central South University, Changsha, China, in 2018, and Ph.D. degree in surveying and mapping science and technology from Central South University, Changsha, China, in 2024. He is currently a postdoctoral with the School of Geography and Remote Sensing, Guangzhou University, Guangzhou, China. His research interests include computer vision, continual learning, and remote sensing image processing.
\end{IEEEbiography}

  \begin{IEEEbiography}
  [{\includegraphics[width=1in,height=1.25in,clip,keepaspectratio]{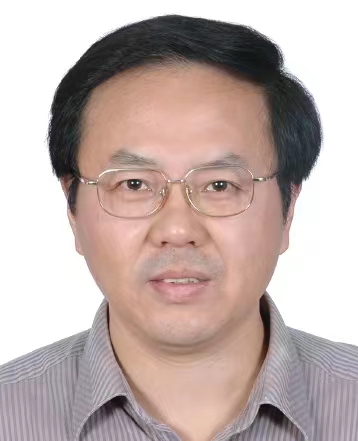}}]{Xinchang Zhang} received the B.S. degree in car-tography from the Wuhan Institute of Surveying and Mapping, Wuhan, China, in 1982, the M.S. degree in cartography from the Wuhan Technical University of Surveying and Mapping, Wuhan, in 1994, and the Ph.D. degree in resources and environmental sciences from Wuhan University, Wuhan, China, in 2004.,He is currently a Professor with the School of Geography and Remote Sensing, Guangzhou University, Guangzhou, China, and a Chair Professor with Henan University, Kaifeng, China. His research interests include spatial database updating, spatial data integration, and smart city.
\end{IEEEbiography}

  \begin{IEEEbiography}
  [{\includegraphics[width=1in,height=1.25in,clip,keepaspectratio]{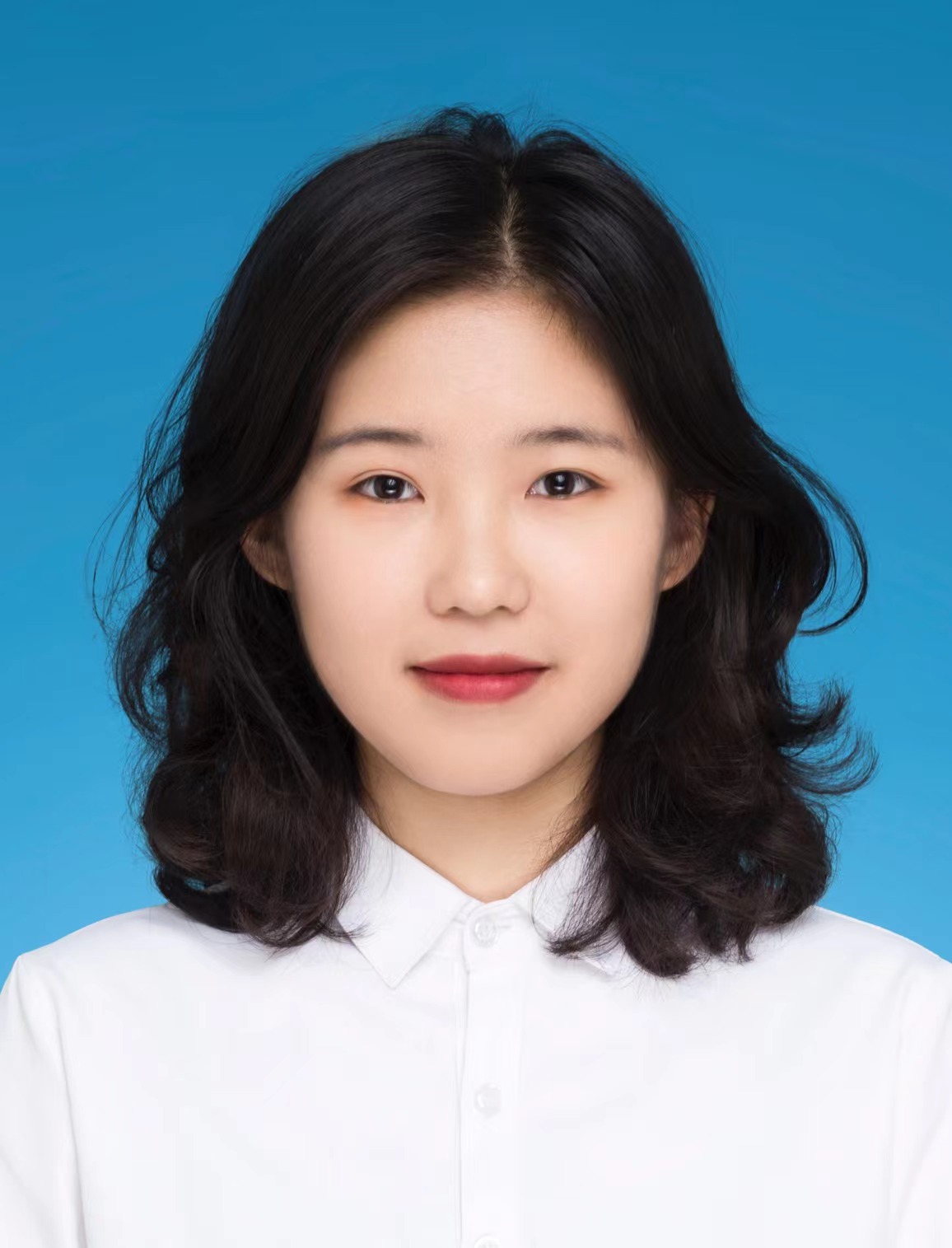}}]{Dingqi Ye} received a B.S. and M.S. from Central South University, Changsha, China, and is currently pursuing an Ph.D. degree in Univeristy of illinois Urbana-Champaign. Her research interests include computer vision, continual learning, and remote sensing image processing.
\end{IEEEbiography}

\begin{IEEEbiography}
  [{\includegraphics[width=1in,height=1.25in,clip,keepaspectratio]{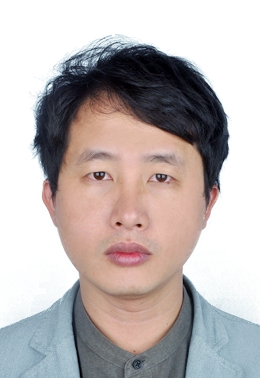}}]{Yongjian Ruan} received the B.S. degree in Geographic Information System from the Jiaying University, Meizhou, China, in 2014, the M.S. degree in Cartography and Geographical Information System from the Jiangxi University of Science and Technology, Ganzhou, China, in 2017, and the Ph.D degree in Cartography and Geographical Information System from Sun Yat-sen University, Guangzhou, China, in 2021. He is currently a Lecturer with the School of Geography and Remote Sensing, Guangzhou University, Guangzhou, China. His research interests include ecological remote sensing, deep learning for extracting remote Sensing information, vegetation phenology and ice phenology. His research has appeared in Journal of Geophysical Research: Atmospheres/biogeosciences, International Journal of Remote Sensing, Giscience and Remote Sensing, Journal of Cleaner Production, Sustainability, among others.
\end{IEEEbiography}

\begin{IEEEbiography}
  [{\includegraphics[width=1in,height=1.25in,clip,keepaspectratio]{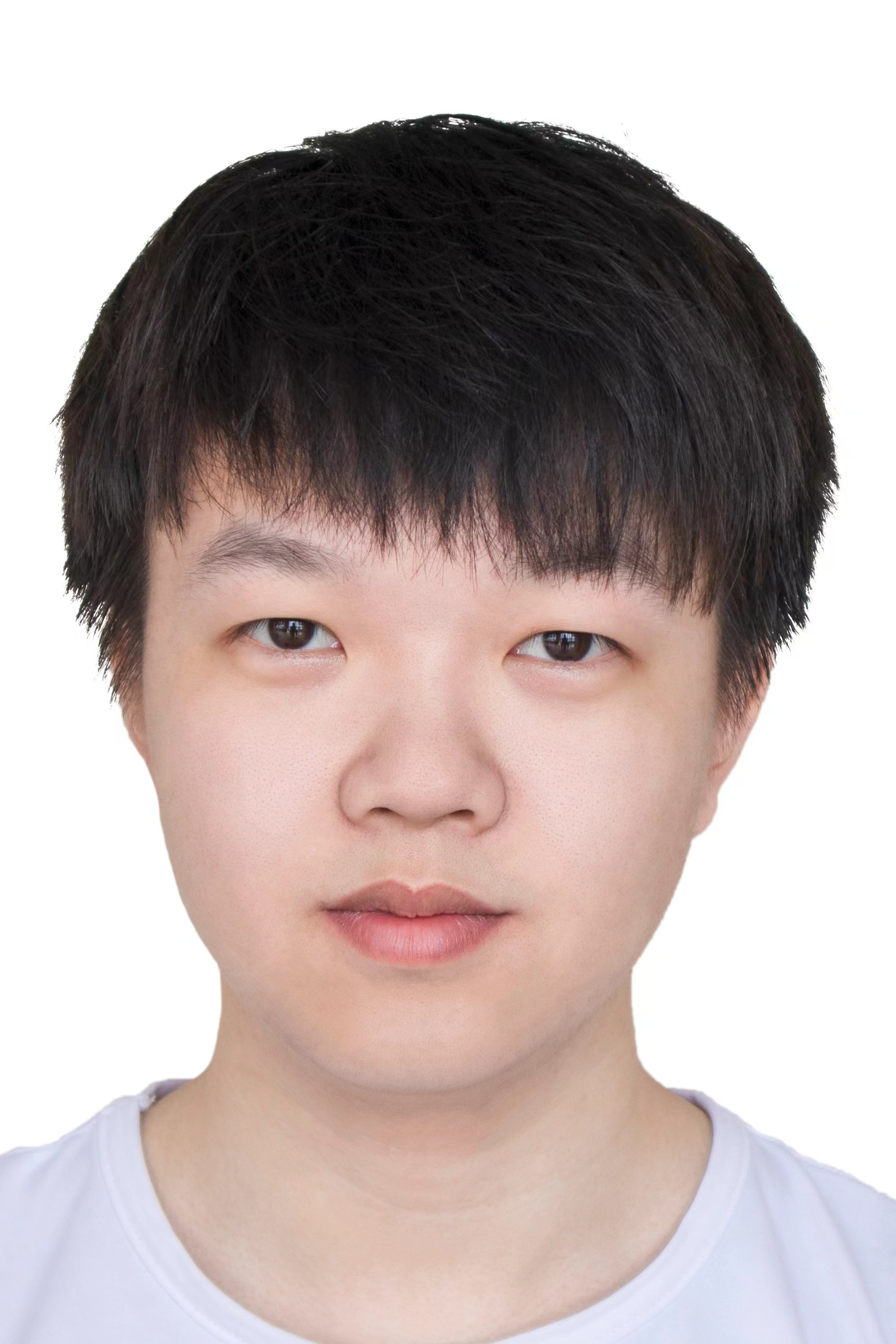}}]{Xin Guo} attained the B.E. degree in remote sensing science and technology from Central South University, Changsha, China, in 2022. Presently, he is actively pursuing his M.Sc. degree within the School of Geosciences and Info-Physics at the same university. His research interests include diffusion models, remote sensing image understanding, and noise label learning.
\end{IEEEbiography}

  \begin{IEEEbiography}
  [{\includegraphics[width=1in,height=1.25in,clip,keepaspectratio]{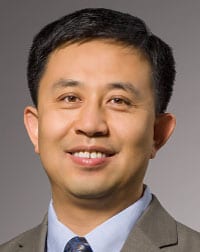}}]{Shaowen Wang}  received the B.S. degree in computer engineering from Tianjin University, Tianjin, China, in 1995, M.S. degree in geography from Peking University, Beijing, China, in 1998, and Master's of Computer Science and the Ph.D. degrees in geography from the University of Iowa, Iowa City, IA, USA, in 2004.,He is a Professor and Head of the Department of Geography and Geographic Information Science; Richard and Margaret Romano Professorial Scholar; and an Affiliate Professor with the Department of Computer Science, Department
  of Urban and Regional Planning, and School of Information Sciences, the University of Illinois at Urbana-Champaign (UIUC), Champaign, IL, USA. He has served as Founding Director of the CyberGIS Center for Advanced Digital and Spatial Studies, UIUC, since 2013. He has authored or coauthored 100+ peer-reviewed papers including articles in 30+ journals. His research interests include geographic information science and systems (GIS), advanced cyberinfrastructure and cyberGIS, complex environmental and geospatial problems, computational and data sciences, high-performance and distributed computing, and spatial analysis and modeling.,Dr. Wang has served as an Action Editor for GeoInformatica, and Guest Editor or editorial board Member for multiple other journals, book series, and proceedings.
\end{IEEEbiography}

  \begin{IEEEbiography}
  [{\includegraphics[width=1in, height=1.25in,clip,keepaspectratio]{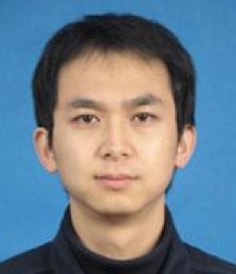}}]{Haifeng Li} received a master's degree in transportation engineering from the South China University of Technology, Guangzhou, China, in 2005, and a Ph.D. degree in photogrammetry and remote sensing from Wuhan University, Wuhan, China, in 2009. He is currently a professor at the School of Geosciences and Info-Physics, Central South University, Changsha, China. He was a research associate with the Department of Land Surveying and Geo-Informatics, The Hong Kong Polytechnic University, Hong Kong, in 2011, and a visiting scholar with the University of Illinois at Urbana-Champaign, Urbana, IL, USA, from 2013 to 2014. He has authored over 30 journal papers. His current research interests include geo/remote sensing big data, machine/deep learning, and artificial/brain-inspired intelligence. He is a reviewer for many journals.
\end{IEEEbiography}

\end{document}